\documentclass[journal]{IEEEtran}
\usepackage{graphicx}
\usepackage{float} 
\usepackage{subfigure}
\usepackage{bm}
\usepackage{amsmath}
\usepackage{amsthm}
\usepackage[ruled]{algorithm2e}
\usepackage{pifont}
\usepackage{color}
\usepackage{amsfonts}
\usepackage{amsmath}
\usepackage{mathrsfs} 
\usepackage{multirow}
\usepackage{balance}
\usepackage{threeparttable} 
\usepackage{epstopdf}
\usepackage{epsfig}
\usepackage{graphicx,booktabs}
\usepackage[pagebackref=true,breaklinks=true,colorlinks,bookmarks=false]{hyperref}
\usepackage{arydshln}
\usepackage{bbm}
\usepackage{threeparttable}
\usepackage{cite}
\usepackage{soul}

\begin{document}
\title{A Parameter-free Nonconvex Low-rank Tensor Completion Model for Spatiotemporal Traffic Data Recovery}

\author{Yang He,
    ~Yuheng~Jia, ~\IEEEmembership{Member,~IEEE,}~Liyang Hu,~Chengchuan An, ~Zhenbo Lu  and Jingxin~Xia
    \thanks{Manuscript received June 22, 2022.
        (Corresponding authors: Jingxin Xia).}
    \thanks{Y. He, C. An, Z. Lu, and J. Xia are with the Intelligent Transportation System Research Center, Southeast University, Nanjing, 211189, China (e-mail: yanghe@seu.edu.cn, ccan@seu.edu.cn, luzhenbo@seu.edu.cn, xiajingxin@seu.edu.cn).}
    \thanks{Y. Jia is with the School of Computer Science and Engineering, Southeast
    University, Nanjing, 211189, China (e-mail: yhjia@seu.edu.cn).}	
    \thanks{L. Hu is with the School of Transportation, Southeast University, Nanjing, 211189, China. (e-mail: huliyang@seu.edu.cn).}
    
		}

\maketitle

\begin{abstract}
    Traffic data chronically suffer from missing and corruption, leading to accuracy and utility reduction in subsequent Intelligent Transportation System (ITS) applications. Noticing the inherent low-rank property of traffic data, numerous studies formulated missing traffic data recovery as a low-rank tensor completion (LRTC) problem. Due to the non-convexity and discreteness of the rank minimization in LRTC, existing methods either replaced rank with convex surrogates that are quite far away from the rank function or approximated rank with nonconvex surrogates involving many parameters. In this study, we proposed a Parameter-Free Non-Convex Tensor Completion model (TC-PFNC) for traffic data recovery, in which a log-based relaxation term was designed to approximate tensor algebraic rank. Moreover, previous studies usually assumed the observations are reliable without any outliers. Therefore, we extended the TC-PFNC to a robust version (RTC-PFNC) by modeling potential traffic data outliers, which can recover the missing value from partial and corrupted observations and remove the anomalies in observations. The numerical solutions of TC-PFNC and RTC-PFNC were elaborated based on the alternating direction multiplier method (ADMM). The extensive experimental results conducted on four real-world traffic data sets demonstrated that the proposed methods outperform other state-of-the-art methods in both missing and corrupted data recovery. The code used in this paper is available at: https://github.com/YoungHe49/T-ITS-PFNC.
\end{abstract}

\IEEEpeerreviewmaketitle

\section{Introduction}

\IEEEPARstart{S}{patiotemporal} traffic data collected from heterogeneous sources (e.g., fixed-sensor, floating car) have facilitated a wide range of applications in Intelligent Transportation Systems (ITSs), such as traffic monitoring and forecasting, advanced traffic control, and route guidance. With the development of traffic sensory technologies, the scale and dimension of traffic data are enlarged. In the meantime, the traffic data also suffer from corruption and even missing due to sensor malfunctioning, communication failure, etc., which inevitably undermines their quality and utility in subsequent ITS applications. Hence, developing an effective approach to accurately and robustly recover traffic data from partial and/or corrupted observations is of great importance for ITS. 

\begin{figure}[h]
    \begin{center}
    \includegraphics[width=3.5in]{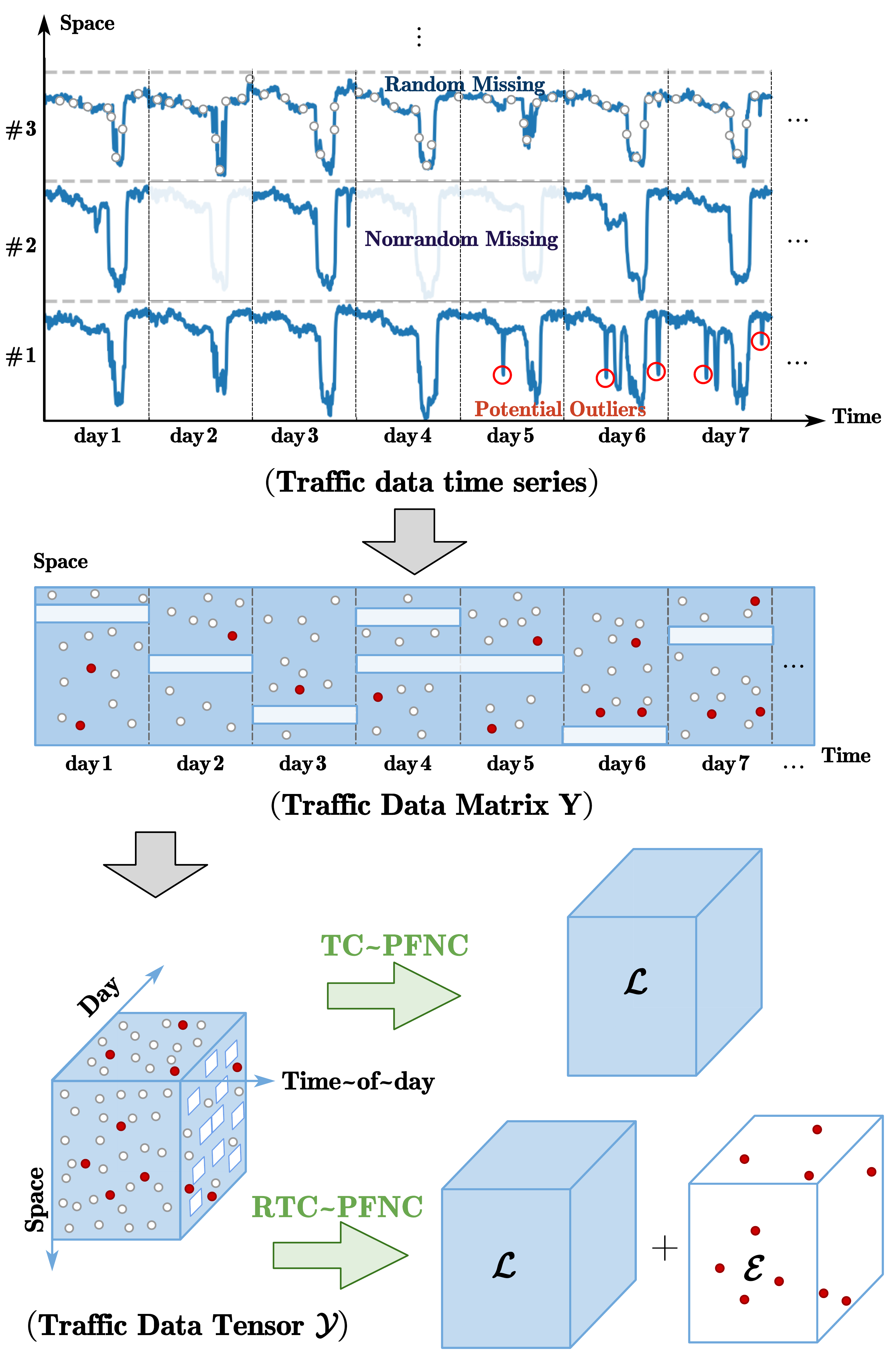}\\
    \caption{Visual illustration of the proposed model. Time series 2 and 3 give the representative examples of traffic data with random and nonrandom missing, respectively, and example 1 displays the potential outlier that existed in collected ``real'' traffic data.}\label{fig1: illustration of traffic data}
    \end{center}
\end{figure}

Traffic data is a high-dimensional time series, which is usually represented as a traffic data matrix (Fig.\ref{fig1: illustration of traffic data}). The traffic data matrix is inherently low-rank, reflected in temporal similarity and spatial correlations between adjacent links and non-adjacent links with similar physical, functional and signal attributes. Therefore, many researchers \cite{wang2018traffic}, \cite{jia2021missing}, \cite{yang2021real}, \cite{chen2021bayesian}, \cite{lei2022bayesian}, \cite{yu2020urban} have attempted to recover the traffic data matrix by utilizing its low-rank nature. To better utilize the traffic data similarity, previous studies further decomposed the temporal dimension into ``time of day" and ``day", and then organized the data into a three-order tensor of size ``space $\times $ time-of-day $\times $ day" as shown in Fig.\ref{fig1: illustration of traffic data}. Correspondingly, the recovery of traffic data can be modeled as a low-rank tensor completion problem (LRTC).

Considering that the rank minimization in LRTC is NP-hard, many researchers solved the convex surrogate \cite{liu2012tensor}, \cite{ran2016tensor}, \cite{chen2021scalable} - nuclear norm minimization (NNM) problem. However, the solutions provided by NNM may seriously deviate from the optimal one because NNM treats all singular values equally and tends to punish the large singular values\cite{nie2018matrix}. Therefore, nonconvex approaches have been proposed and provided great advantages over the convex nuclear norm.

The existing nonconvex low-rank tensor representation approaches can be grouped into two categories: Factorization-based (FB) and Rank Minimization based (RMB). Given a partially observed or corrupted traffic data tensor, the FB methods aim to find a low-rank tensor that is consistent with the observed one on known entries by decomposing it into smaller factor tensors with pre-defined rank. Typical FB methods include: tensor CP factorization\cite{chen2021bayesian}, \cite{asif2016matrix}, \cite{chen2019bayesian}, \cite{chen2019missing}, \cite{baggag2019learning}, tensor Tucker factorization\cite{tan2013tensor}, \cite{tan2016short}, \cite{chen2018spatial}, \cite{zhang2019missing}, tensor SVD factorization\cite{deng2021graph}, \cite{feng2022traffic}, and tensor train factorization\cite{zhang2021tensor}. Although those methods have achieved great success in traffic data recovery, the rank of decomposed tensor must be pre-defined, which is difficult in real situations.

Instead of decomposing the observed tensor into factor tensors, the RMB methods aim to directly recover the low-rank tensor. The existing nonconvex RMB methods include: Truncate Nuclear Norm (TNN) minimization \cite{chen2020nonconvex},\cite{chen2021low}, and Schatten-$p$ norm minimization \cite{nie2022truncated}. The exact structure information (e.g. pre-defined rank) of the recovered tensor is not required in RMB methods. Nevertheless, they still need additional parameters to control the model non-convexity, such as truncation rate $r$ in TNN \cite{chen2020nonconvex}, \cite{chen2021low}, and value $p$ in Schatten-$p$ norm minimization \cite{nie2022truncated}. In order to choose a proper value for the additional parameters, parameter fine-tuning and try-and-error are usually required in large-scale real-world practice where the hidden patterns in spatiotemporal traffic data are unknown.



In summary, the low-rank tensor representation models have achieved great success in the field of traffic data recovery, of which nonconvex approaches are more accurate than convex methods. However, labour-intense parameter works are still required for the nonconvex approaches, such as the pre-estimation of rank in FB methods and the calibration of additional non-convexity control parameters in RMB methods. Additionally, there has been less evidence for modeling potential traffic data outliers in previous nonconvex approaches. 

In this paper, we proposed a parameter-free nonconvex regularizer and utilized it to construct two low-rank tensor completion models, aiming to improve the \textbf{precision}, \textbf{applicability}, and \textbf{robustness} of traffic data recovery. The contributions and findings of this work are summarized as follows:

\begin{enumerate}
\item We first presented a log-based nonconvex regularizer to approximate tensor algebraic rank, which can also simultaneously increase the punishment on noise and decrease the punishment on structural information. Specially, the regularizer does not involve any parameter. Then, we utilized it to construct a tensor completion model for traffic data recovery, named TC-PFNC (Tensor Completion based on Parameter-Free Non-Convex relaxation);
\item Considering that the observed traffic data may be corrupted by outliers, we extended the TC-PFNC to RTC-PFNC (Robust TC-PFNC) to recover missing values from partial and corrupted observations and remove the anomalies in observations synchronously;
\item The effectiveness and superiority of the proposed method were proved by extensive numerical experiments on four real-world traffic data sets under missing data and corrupted and missing data settings. 
\end{enumerate}

The rest of this paper is organized as follows. In Section \ref{sec: Related Work}, we briefly reviewed existing nonconvex low-rank approximation methods for traffic data recovery. We presented the proposed parameter-free nonconvex tensor completion model (TC-PFNC) and its robust extension (RTC-PFNC) and their numerical solution in Section \ref{sec:Methodology}, followed by traffic data sets and baseline models description in Section \ref{sec:Experiment} as well as extensive experimental results and analysis on missing data and corrupted missing data scenarios in Section \ref{sec:Results}. Finally, Section \ref{sec:Conclusion} concluded this paper. The notations used in this paper are presented in Tab. \ref{tab:notation}.

\begin{table}[h]
    \centering    
    \caption{Notations.}
    \label{tab:notation}
     {\renewcommand\baselinestretch{1.4}\selectfont 
     \begin{tabular}{l}
    \hline\hline
    1. $x$: scalars. \\
    2. $\boldsymbol{x}$: vectors.\\
    3. $\mathbf{X}\in \mathbb{R} ^{n_1\times n_2}$: matrix with size $n_1\times n_2$.\\
    4. $\mathcal{X} \in \mathbb{R} ^{n_1\times n_2\times n_3}$: 3-order tensor with size $n_1\times n_2\times n_3$.\\
    5. $\mathbf{X}_{\left( k \right)}\in \mathbb{R} ^{n_k\times \frac{n_1\times n_2\times n_3}{n_k}}$: k-mode unfolding matrix of tensor $\mathcal{X}$.\\
    6. $\tilde{\mathcal{X}}\in \mathbb{R} ^{3\times n_1\times n_2\times n_3}$: 4-order tensor with size $3\times n_1\times n_2\times n_3$.\\
    7. $\sigma _i\left( \mathbf{X} \right)$: $i$ th singular value of matrix $\boldsymbol{x}$.\\
    8. $\left\| \mathcal{X} \right\| _F$: frobenius norm of tensor $\mathcal{X}$.\\
    \hline\hline 
    \end{tabular}
    \par}
\end{table}

\section{Related Work}
\label{sec: Related Work}

Given a high-dimensional traffic data time series collected from $n_1$ locations at $t$ time intervals, an intuitive idea is to organize it as a matrix $\mathbf{X}\in \mathbb{R} ^{n_1\times t}$. To better utilize the day-of-week similarity in traffic data, we represent the traffic data matrix $\mathbf{X}\in \mathbb{R} ^{n_1\times t}$ to a three-order tensor $\mathcal{X} \in \mathbb{R} ^{n_1 \times n_2\times n_3}$ by dividing the $t$ time intervals into $n_2$ time intervals of a day and $n_3$ days. To this end, the traffic data recovery problem can be transformed to the completion of low-rank tensor $\mathcal{L}$ based on partially observed traffic data tensor $\mathcal{Y}$ . The general formulation of low-rank tensor completion based on rank minimization is expressed as
\begin{align}
    \min  rank\left( \mathcal{L} \right),
	~~s.t. ~P_{\Omega}\left( \mathcal{L} \right) = P_{\Omega}\left( \mathcal{Y} \right),
	\label{eq:problem2.1}
\end{align}
where $rank$ is the tensor algebraic rank of traffic data tensor $\mathcal{L} \in \mathbb{R} ^{n_1\times n_2\times n_3}$, $\Omega $ denotes the index set of the observed entries, and the operator $P_{\Omega}$ represents the orthogonal projection supported on $\Omega$, i.e.,
\begin{align}
    \left[ P_{\Omega}\left( \mathcal{X} \right) \right] _{ijk}=\begin{cases}
        x_{ijk},~   &\mathrm{if}\left( i,j,k \right) \in \Omega ,\\
        0,      &\mathrm{otherwise}.\\
    \end{cases}
\label{eq:problem2.2}
\end{align}

\begin{figure}[h]
    \begin{center}
    \includegraphics[width=3.5in]{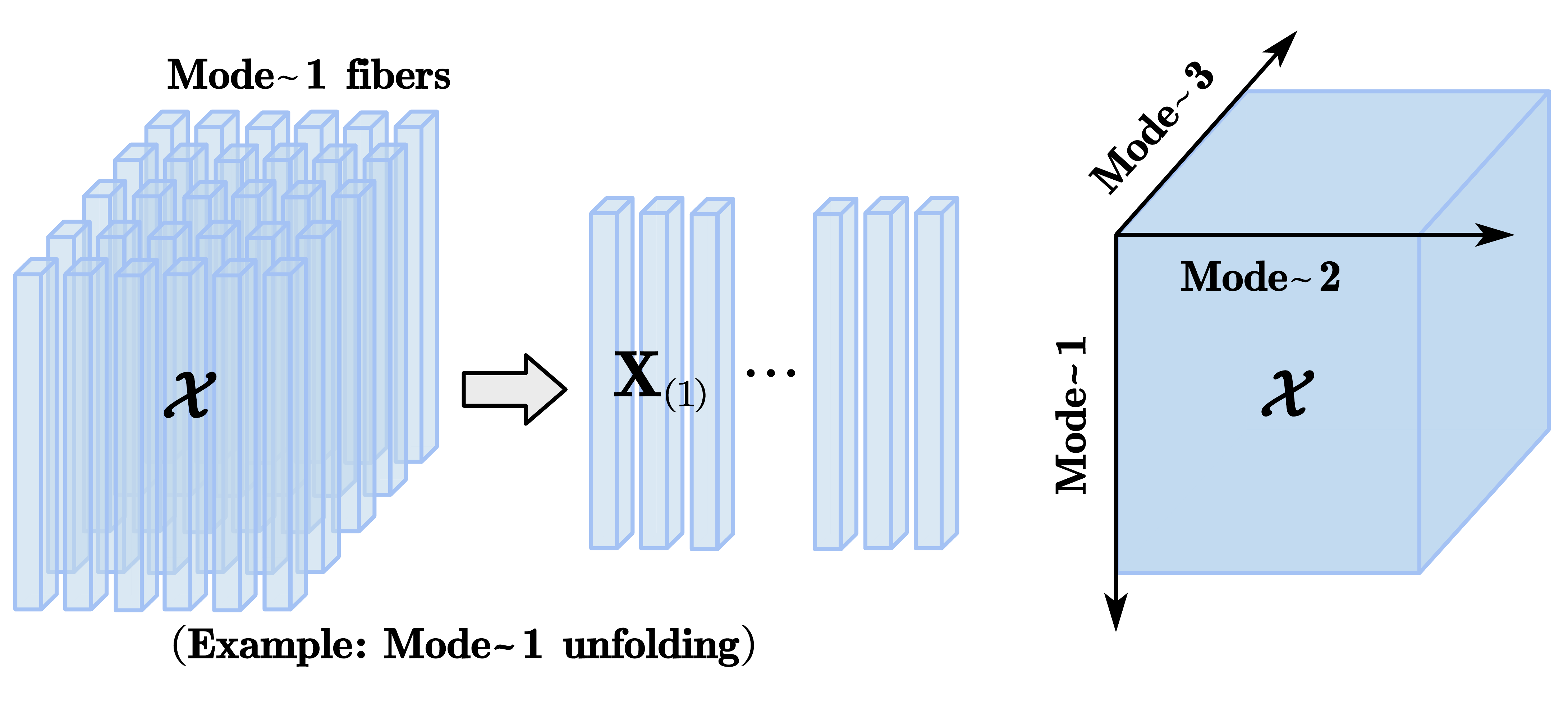}\\
    \caption{Mode-$n$ unfoldings of a three-order tensor.}\label{fig2: illustration of unfolding}
    \end{center}
\end{figure}

The rank minimization problem in Eq.\eqref{eq:problem2.1} is generally NP-hard and computationally intractable. Unlike the matrix rank, the definition of a tensor rank is not unique\cite{jia2021multi}. Numerous alternatives to tensor rank have been proposed, among which a classical approach is to approximate tensor rank using the weighted sum of the convex nuclear norm (SNN)\cite{liu2012tensor} of three unfolded matrices. The rule of three-order tensor unfoldings is illustrated in Fig.\ref{fig2: illustration of unfolding}. To this end, the tensor rank minimization problem in Eq.\eqref{eq:problem2.1} can be rewritten as
\begin{align}
    \underset{\mathcal{L}}{\min}\sum_{k=1}^3{\alpha _k\left\| \mathbf{L}_{k\left( k \right)} \right\| _*},~~s.t.~P_{\Omega}\left( \mathcal{L} \right) =P_{\Omega}\left( \mathcal{Y} \right) ,
	\label{eq:problem2.3}
\end{align}
where $\mathbf{L}_{k\left( k \right)}\in \mathbb{R} ^{n_k\times \left( \prod\nolimits_{i\ne k}^{}{n_i} \right)}$ denotes $k$-th unfolded matrix of tensor $\mathcal{L} _k$,  $\alpha _k$ is the weight for $\mathbf{L}_{k\left( k \right)}$, and $\left\| \cdot \right\| _*$ indicates the nuclear norm regularizer.

Based on SNN, Chen \textit{et al}.\cite{chen2020nonconvex} replaced nuclear norm regularizar in Eq.\eqref{eq:problem2.3} with nonconvex Truncated Nuclear Norm (TNN) and reformulated the low-rank tensor completion as
\begin{align}
    \underset{\mathcal{L}}{\min}\sum_{k=1}^3{\alpha _k\left\| \mathbf{L}_{k\left( k \right)} \right\| _{r_k,*}},~~s.t.~P_{\Omega}\left( \mathcal{L} \right) =P_{\Omega}\left( \mathcal{Y} \right),
\label{eq:problem2.4}
\end{align}
where the TNN regularizer $\left\| \cdot \right\| _{r_k,*}$ is defined as the sum of $\min \left\{ m,n \right\} -r_k$ minimum singular values of matrix $\mathbf{X}\in \mathbb{R} ^{m\times n}$, i.e. 
\begin{align}
    \left\| \mathbf{X} \right\| _{r_k,*}=\sum\nolimits_{i=r_k+1}^{\min \left\{ m,n \right\}}{\sigma _i\left( \mathbf{X} \right)},  
\label{eq:problem2.5}
\end{align}
with the parameter $r_k$ to control the degree of truncation.

Yu \textit{et al}.\cite{yu2020urban} proposed a noncovex regularizer $\left\| \cdot \right\| _{S_p}$ based on Schatten $p$-norm for traffic speed matrix completion as follow
\begin{align}
    \left\| \mathbf{X} \right\| _{S_p}=\left( \sum\nolimits_{i=1}^{\min \left\{ m,n \right\}}{\sigma _i\left( \mathbf{X} \right)}^p \right) ^{\frac{1}{p}}, 
\label{eq:problem2.6}
\end{align}
where $p \in (0,1]$ controls the tightness to algebraic rank, noting that $\left\| \cdot \right\| _{S_p}$ degrades to the nuclear norm if $p=1$.

Taking both advantages of the truncated nuclear norm and Schatten $p$-norm, Nie \textit{et al}.\cite{nie2022truncated} proposed a truncated tensor Schatten $p$-norm model, i.e.,
\begin{align}
    \underset{\mathcal{L}}{\min}\sum_{k=1}^3{\alpha _k\left\| \mathbf{L}_{k\left( k \right)} \right\| _{\theta ,S_p}^{p}},~~s.t.~P_{\Omega}\left( \mathcal{L} \right) =P_{\Omega}\left( \mathcal{Y} \right) , 
\label{eq:problem2.7}
\end{align}
where the truncated Schatten $p$-norm regularizer $\left\| \cdot \right\| _{\theta ,S_p}$ is defined as
\begin{align}
    \left\| \mathbf{X} \right\| _{\theta ,S_p}=\left( \sum\nolimits_{i=r_k+1}^{\min \left\{ m,n \right\}}{\sigma _{i}^{p}\left( \mathbf{X} \right)} \right) ^{\frac{1}{p}}.  
    \label{eq:problem2.8}
\end{align}

Two major insights implicated from the above non-convex relaxations \cite{chen2020nonconvex},\cite{nie2022truncated} include: 1) larger singular values convey the primary information (e.g. periodic and major trends of traffic flow), which should not be punished excessively; 2) smaller singular values commonly represent the noisy information (e.g. irregular fluctuations), which should be punished as zeros.




\section{Methodology}
\label{sec:Methodology}

\subsection {Low-rank Tensor Completion with Parameter-Free Non-Convex relaxation (TC-PFNC)}
\label{A: TC-PFNC}
\begin{figure}[t]
    \begin{center}
    \includegraphics[width=2.5in]{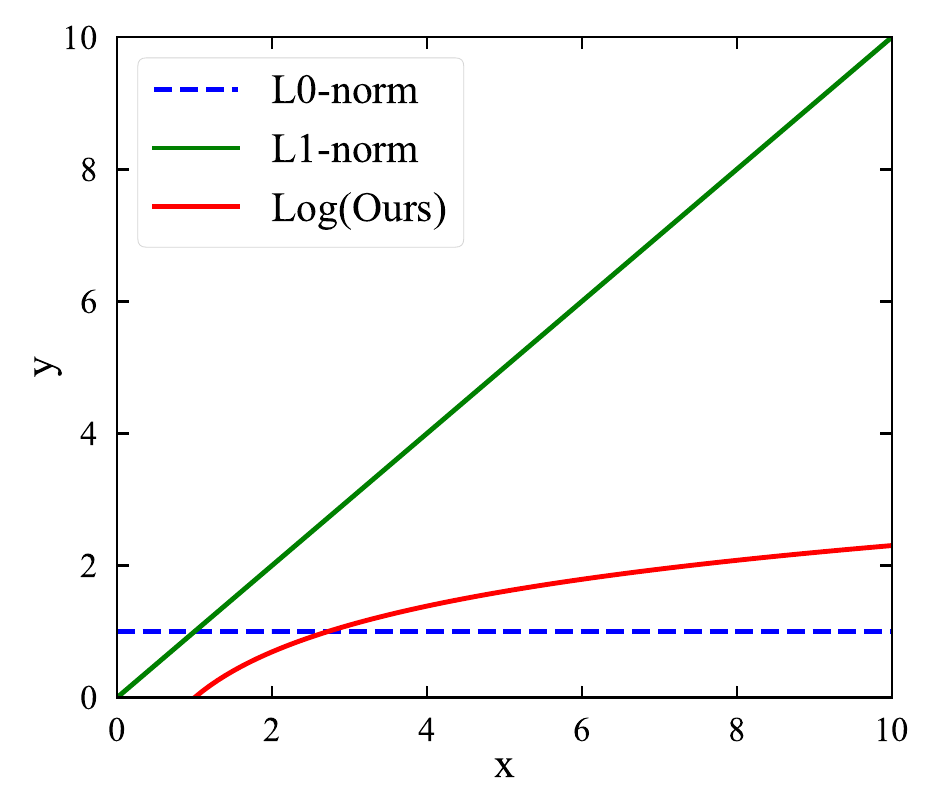}
    \caption{One dimensional illustrations with respect to three regularization terms.}
    \label{fig2: Log}
    \end{center}
\end{figure}

Previous works have validated the effectiveness and advantages of exploring the low-rankness of traffic data by using nonconvex relaxation. However, additional parameters are required in these methods, which may limit their applicability to real-world practice. To this end, we introduced a parameter-free non-convex regularizer
\begin{align}
    f\left( \mathbf{X} \right) =\sum_{i=1}^n{log\left( \sigma _i\left( \mathbf{X} \right) +\varepsilon \right)}
\label{eq:problem3.1.1.1}
\end{align}
with a small constant $\varepsilon>0$ to ensure positivity, and recasted the low-rank tensor completion problem as
\begin{align}
    \underset{\mathcal{L}}{\min}\sum_{k=1}^3{\alpha _kf\left( \mathbf{L}_{k\left( k \right)} \right)},~~s.t.~ P_{\Omega}\left( \mathcal{L} \right) =P_{\Omega}\left( \mathcal{Y} \right) ,
\label{eq:problem3.1.1.2}
\end{align}
where $\mathbf{L}_{k\left( k \right)}\in \mathbb{R} ^{n_k\times \left( \prod\nolimits_{i\ne k}^{}{n_i} \right)}$ denotes $k$-th unfolded matrix of tensor $\mathcal{L} _k$.  As illustrated in Fig.\ref{fig2: Log}, we can observe that $f\left( \mathbf{X} \right)$ is a neutralization between rank minimization (the $l_0$-norm) and nuclear norm (the $l_1$-norm) minimization, which can simultaneously increase the punishment on smaller singular values (i.e. noise) and decrease the punishment on larger singular values (i.e. primary information).

We introduced auxiliary variable $\mathcal{M}$ to keep observation information from observed tensor $\mathcal{Y}$  and then broadcast it to three variables $\mathcal{L}_1,\mathcal{L}_2,\mathcal{L}_3$ that correspond to the three unfolded matrices of $\mathcal{L}$, the TC-PFNC model in Eq.\eqref{eq:problem3.1.1.2} is equivalently rewritten as
\begin{align}
    \underset{\mathcal{L} _1,\mathcal{L} _2,\mathcal{L} _3,\mathcal{M}}{\min}&\sum_{k=1}^3{\alpha _kf\left( \mathbf{L}_{k\left( k \right)} \right)}, \nonumber \\
    s.t.~\mathcal{L} _k=\mathcal{M} ,k=1,&2,3,P_{\Omega}\left( \mathcal{M} \right) =P_{\Omega}\left( \mathcal{Y} \right) ,
\label{eq:problem3.1.2}
\end{align}

To obtain the optimal solution of the proposed model, we adopted the widely used Alternating Direction Method of Multipliers (ADMM) framework to decompose the problem in Eq. \eqref{eq:problem3.1.2} into several easy-to-handle subproblems. To cope with the equality constraints, the augmented Lagrangian function of our TC-PFNC model was rewritten as
\begin{align}
    &\underset{\mathcal{L} _1,\mathcal{L} _2,\mathcal{L} _3,\mathcal{M}}{\mathrm{arg}\min}\sum_{k=1}^3{\Big\{}\alpha _kf\left( \mathbf{L}_{k\left( k \right)} \right) +\frac{\rho _k}{2}\left\| \mathcal{L} _k-\mathcal{M} \right\| _{F}^{2} \nonumber\\
    &~~~~~~~~~+\left< \mathcal{T} _{k},\mathcal{L} _k-\mathcal{M} \right> \Big\}, ~s.t.~P_{\Omega}\left( \mathcal{M} \right) =P_{\Omega}\left( \mathcal{Y} \right) ,
    \label{eq:problem3.1.3}
\end{align}
where $\left< \cdot ,\cdot \right>$ indicates the inner product, $\mathcal{T}_k\in \mathbb{R} ^{n_1\times n_2\times n_3}$ denotes the Lagrangian multiplier and $\rho_k>0$ represents the penalty parameter of $k$-th mode. According to the ADMM framework, the minimization of the TC-PFNC
model can be decomposed into iteratively solving the following three subproblems:
\begin{align}
    \mathcal{L} _{k}^{l+1}&:=\underset{\mathcal{L}_k}{\mathrm{arg}\min}\sum_{k=1}^3{\Big\{}\alpha _kf\left( \mathbf{L}_{k\left( k \right)} \right)\nonumber \\
    &~~~~~~~~~~~~~+\frac{\rho _k}{2}\left\| \mathcal{L} _k-\mathcal{M}^l \right\| _{F}^{2}+\left< \mathcal{T} _k^l,\mathcal{L} _k \right> \Big\},
    \label{eq:problem3.1.4}
\end{align}
\begin{align}
    \mathcal{M} ^{l+1}:=\underset{\mathcal{M}}{\mathrm{arg}\min}&\sum\nolimits_{k=1}^3{\Big\{}\left< \mathcal{T} _{k}^{l},-\mathcal{M} \right> \nonumber\\
    &~~~~~~~~~~+\frac{\rho _k}{2}\left\| \mathcal{L} _{k}^{l+1}-\mathcal{M} \right\| _{\mathrm{F}}^{2}\Big\},
    \label{eq:problem3.1.5}
\end{align}
\begin{align}
    &\mathcal{T} _{k}^{l+1} :=\mathcal{T} _{k}^{l}+\rho _k\left( \mathcal{L} _{k}^{l+1}-\mathcal{M} ^{l+1} \right), ~~~~~~~~~~~~~~~~~
 \label{eq:problem3.1.6}
\end{align}
where $l$ denotes the $l$-th iteration, and the three types of variables $\mathcal{L} _k, \mathcal{M},\mathcal{T} _k $ are alternatively updated in each iteration (i.e., updating one with the others fixed) until convergence. The detailed solutions of Eq.\eqref{eq:problem3.1.4} and Eq.\eqref{eq:problem3.1.5} are introduced in the next subsections. The pseudocode of TC-PFNC is given in Algorithm \ref{algorithm1: TC-PFNC}.

\subsubsection{Update variable $\mathcal{L}$} 
\label{update L}
To cope with the form of $\mathbf{L}_{k\left( k \right)}$, we unfolded all the tensor along the $k$-th mode and convert Eq.\eqref{eq:problem3.1.4} as follows
\begin{align}
    \mathcal{L} _{k}^{l+1}&=\underset{\mathcal{L} _k}{\mathrm{arg}\min}\Big\{ \alpha _kf\left( \mathbf{L}_{k\left( k \right)} \right) \nonumber\\
    &~~~+\frac{\rho _k}{2}\left\| \mathbf{L}_{k\left( k \right)}-\mathbf{M}_{k\left( k \right)}^{l} \right\| _{\mathrm{F}}^{2}+\left< \mathbf{T}_{k\left( k \right)}^{l}, \mathbf{L}_{k\left( k \right)}\right> \Big\} \nonumber\\
    &=\underset{\mathcal{L} _k}{\mathrm{arg}\min}\Big\{  \frac{\alpha _k}{\rho _k}f\left( \mathbf{L}_{k\left( k \right)} \right) \nonumber\\
    &~~~+\frac{1}{2}\left\| \mathbf{L}_{k\left( k \right)}-\left( \mathbf{M}_{k\left( k \right)}^{l}-\frac{1}{\rho _k}\mathbf{T}_{k\left( k \right)}^{l} \right) \right\| _{\mathrm{F}}^{2}  \Big\} \nonumber\\
    &=\mathrm{fold}_k\left( \mathcal{D} _{\omega ^l,\tau}\left( \mathbf{M}_{k\left( k \right)}^{l}-\frac{1}{\rho _k}\mathbf{T}_{k\left( k \right)}^{l} \right) \right),  
    \label{eq:problem3.1.7}
\end{align}
where $\mathcal{D} _{\boldsymbol{\omega }^{\boldsymbol{l}},\tau}\left( \cdot \right)$ denotes the weighted singular value thresholding operator as shown in Lemma 1. 

\textbf{Lemma.1}\cite{dong2014compressive}. Given $\tau >0$, $\mathbf{Z}\in \mathbb{R} ^{m\times n}$, an optimal solution to the problem
\begin{align}
    \underset{X}{\min}\frac{1}{2}\left\| \mathbf{X}-\mathbf{Z} \right\| _{\mathrm{F}}^{2}+\tau f\left( \mathbf{X} \right) ,
\label{eq:problem3.1.8}
\end{align}
is given by the weighted singular value thresholding operator:
\begin{align}   
    \mathcal{D} _{\boldsymbol{\omega }^{\boldsymbol{l}},\tau}\left( \mathbf{Z} \right) =\mathbf{U}\left( \mathbf{\Sigma }-\tau \mathrm{diag}\left( \boldsymbol{\omega }_{\boldsymbol{l}} \right) \right) _+\mathbf{V}^{\top},
\label{eq:problem3.1.9}
\end{align}
where $\boldsymbol{U}\left( \boldsymbol{\varSigma } \right) \boldsymbol{V}^{\top}$  is the singular value decomposition of $\mathbf{Z}$ , the $\boldsymbol{\omega }^{\boldsymbol{l}}$  is defined as 
\begin{align}
    \boldsymbol{\omega }^{\boldsymbol{l}}=\left( \frac{1}{\sigma _1\left( \mathbf{X}^l \right) +\varepsilon _k},...,\frac{1}{\sigma _{\min \left\{ m,n \right\}}\left( \mathbf{X}^l \right) +\varepsilon _k} \right),
 \label{eq:problem3.1.10}
\end{align}
where $\sigma _{1}^{}\left( \mathbf{X}^l \right) ,...,\sigma _{\min \left\{ m,n \right\}}^{}\left(\mathbf{X}^l \right) $ are the solutions obtained in the $l$th iteration.

Gathering the results of $\mathcal{L}_1,\mathcal{L}_2,\mathcal{L}_3$ in Eq.\eqref{eq:problem3.1.7}, we can update the variable $\mathcal{L}$ by
\begin{align}
    \mathcal{L} ^{l+1}=\sum_{k=1}^3{\alpha _k\mathcal{L} ^{l+1}_k}.
\label{eq:problem3.1.11}
\end{align}

\subsubsection{Update variable $\mathcal{M}$}
\label{update M}
Specifically, the $\mathcal{M}$ sub-problem in \eqref{eq:problem3.1.5} is a set of unconstrained quadratic equations elementwise. Therefore, the closed-form solution is obtained as
\begin{align}
    \mathcal{M} ^{l+1}&= \underset{\mathcal{M}}{\mathrm{arg}\min}\left< \mathcal{M} , \mathcal{M} \right> \nonumber\\
    &~~~-\frac{2}{\sum\nolimits_{k=1}^3{\rho _k}}\left< \mathcal{M} , \sum\nolimits_{k=1}^3{\left( \rho _k\mathcal{L} _{k}^{l+1}+\mathcal{T} _{k}^{l} \right)} \right> \nonumber\\
    &= \frac{1}{\sum\nolimits_{k=1}^3{\rho _k}}\sum\nolimits_{k=1}^3{\left( \rho _k\mathcal{L} _{k}^{l+1}+\mathcal{T} _{k}^{l} \right)},
\label{eq:problem3.1.12}
\end{align}
where we imposed a fixed constraint, i.e. $P_{\Omega}\left( \mathcal{M} _{\mathbf{\Omega }}^{l+1} \right) =P_{\Omega}\left( \mathcal{Y} \right) $, to guarantee the transformation of observation information at each iteration.


\subsubsection{Update variable $\mathcal{T}$}
In practice, if $\rho _1=\rho _2=\rho _3=\rho $ , the variable  $\mathcal{T} _{k}^{l+1}$ can be updated by
 \begin{align}
    \tilde{\mathcal{T}}^{l+1}:=\tilde{\mathcal{T}}^l+\rho \left( \tilde{\mathcal{L}}^{l+1}-\tilde{\mathcal{M}}^{l+1} \right),
 \label{eq:problem3.1.13}
\end{align}
where $\tilde{\mathcal{T}},\tilde{\mathcal{L}},\tilde{\mathcal{E}},\tilde{\mathcal{M}}$ are fourth-order tensors of the size $3\times M\times N\times T$. Specifically, $\tilde{\mathcal{T}}, \tilde{\mathcal{L}}, \tilde{\mathcal{E}}$ are stacked by three third-order tensors $\mathcal{T} _k, \mathcal{L} _k, \mathcal{E} _k$  over the fourth mode, respectively, and $\tilde{\mathcal{M}}$ is stacked over the fourth mode by copying the third-order tensor $\mathcal{M}$.

\begin{algorithm}[h]
    \caption{Numerical solution of \eqref{eq:problem3.1.3} via ADMM}
    \KwIn{The Observed Data Tensor $\mathcal{Y}$}
    \KwOut{The Recovered Low-rank Tensor $\mathcal{L}$}
    \textbf{Initialization:} $\mathcal{L} ^0=\mathcal{M}^0 =\mathcal{Y},~\mathcal{L}_k ^0=\mathcal{T}_k ^0=0,\alpha _1=\alpha _2=\alpha _3=\frac{1}{3},~\rho_k=\rho =\rho ^0,~\varepsilon =1\mathrm{e}-6$;
    \LinesNumbered
    
    \SetAlgoVlined
    \While{not converged}{
    \For{k=1:3}{
        Update $\mathcal{L} _{k}^{l+1}$ via Eq.\eqref{eq:problem3.1.7} \;
        }
    Update $\mathcal{L} ^{l+1}$ via Eq.\eqref{eq:problem3.1.11}\;
    Update $\mathcal{M} ^{l+1}$ via Eq.\eqref{eq:problem3.1.12}\;
    Update $\tilde{\mathcal{T}}^{l+1}$ via Eq. \eqref{eq:problem3.1.13}\;
    $l=l+1$
    }
\label{algorithm1: TC-PFNC}
\end{algorithm}

\subsection {Robust Extension (RTC-PFNC)}
\label{B: RTC-PFNC}
Note that the TC-PFNC is constructed for solving the noiseless tensor completion problem, while in real practice there exist many outliers in traffic data observations, such as time series 1 in fig.\ref{fig1: illustration of traffic data}. In this section, in order to eliminate the nagative effects of outliers on traffic data recovery, we extended TC-PFNC to a robust version (RTC-PFNC) by introducing a anomaly term $\mathcal{E}$ and reformulated the low-rank tensor completion model in Eq.\eqref{eq:problem3.1.2} as
 \begin{align}
    \underset{\left\{ \mathcal{L} _k,\mathcal{E} _k \right\} _{k=1}^{3},\mathcal{M}}{\min}&\sum_{k=1}^3{\alpha _kf\left( {\mathbf{L}_k}_{\left( k \right)} \right)}+\lambda _k\left\| \mathcal{E} _k \right\| _1, \nonumber\\
    s.t.~\mathcal{L} _k&+\mathcal{E} _k=\mathcal{M},P_{\Omega}\left( \mathcal{M} \right) =P_{\Omega}\left( \mathcal{Y} \right), 
\label{eq:problem3.2.1}
\end{align}
where $\left\| \cdot \right\| _1$ denotes the $l_1$-norm of tensors, $\mathcal{E}_k$ is the anomaly term, and $\lambda_k$ is the weight for $\mathcal{E}_k$.
The augmented Lagrangian function of Eq.\eqref{eq:problem3.2.1} is writtern as
\begin{align}
    &\mathop {\mathrm{arg}\min} \limits_{\left\{ \mathcal{L} _k,\mathcal{E} _k, \right\} _{k=1}^{3},\mathcal{M}}\sum_{k=1}^3{\Big\{}\alpha _kf\left( \mathbf{L}_{k\left( k \right)} \right) +\lambda _k\left\| \mathcal{E} _k \right\| _1 \nonumber\\
    &~~~~+\frac{\rho _k}{2}\left\| \mathcal{L} _k+\mathcal{E} _k-\mathcal{M} \right\| _{F}^{2}+\left< \mathcal{T} _k,\mathcal{L} _k+\mathcal{E} _k-\mathcal{M} \right> \Big\},\nonumber\\
    &~s.t.~P_{\Omega}\left( \mathcal{M} \right) =P_{\Omega}\left( \mathcal{Y} \right) ,
 \label{eq:problem3.2.2}
\end{align}
where $\mathcal{T}_k\in \mathbb{R} ^{n_1\times n_2\times n_3}$ denotes the Lagrangian multiplier  used for dual update in the following ADMM scheme. Accordingly, the ADMM transforms the robust tensor completion (RTC-PFNC) to the following subproblems in an iterative manner:
\begin{align}
    \mathcal{L} _{k}^{l+1}:=&\mathop {\mathrm{arg}\min} \limits_{\mathcal{L} _k}\sum_{k=1}^3{\Big\{}\alpha _kf\left( \mathbf{L}_{k\left( k \right)} \right)+\left< \mathcal{T} _{k}^{l},\mathcal{L} _k \right> \nonumber\\
    &~~~~~~~~~~~~~~~~+\frac{\rho _k}{2}\left\| \mathcal{L} _k+\mathcal{E} _{k}^{l}-\mathcal{M} ^l \right\| _{F}^{2} \Big\},~~~
    \label{eq:problem3.2.3}
\end{align}
\begin{align}
    \mathcal{M} ^{l+1}:=&\mathop {\mathrm{arg}\min} \limits_{\mathcal{L} _k}\sum_{k=1}^3\Big\{\left< \mathcal{T} _{k}^{l},-\mathcal{M} \right> \nonumber\\
    &~~~~~~~~~~~~+ \frac{\rho _k}{2}\left\| \mathcal{L} _{k}^{l}+\mathcal{E} _{k}^{l}-\mathcal{M} \right\| _{F}^{2} \Big\},~~~~~
    \label{eq:problem3.2.4}
\end{align}
\begin{align}
    \mathcal{E} _{k}^{l+1}:=&\mathop {\mathrm{arg}\min} \limits_{\mathcal{E} _k}\Big\{ \lambda _k\left\| \mathcal{E} _k \right\| _1+\left< \mathcal{T} _{k}^{l},\mathcal{E} _k \right> \nonumber \\
    &~~~~~~~~~~~~~+\frac{\rho _k}{2}\left\| \mathcal{L} _{k}^{l}+\mathcal{E} _k-\mathcal{M} ^l \right\| _{F}^{2}\Big\}, ~~~~~
    \label{eq:problem3.2.5}
\end{align}
\begin{align}
    \tilde{\mathcal{T}}^{l+1}=\tilde{\mathcal{T}}^l+\rho \left( \tilde{\mathcal{L}}^{l+1}+\tilde{\mathcal{E}}^{l+1}-\tilde{\mathcal{M}}^{l+1} \right),~~~~~~~~~~~~
    \label{eq:problem3.2.6}
\end{align}
where $l$ denotes the $l$-th iteration. The detailed solutions of Eq.\eqref{eq:problem3.2.3}, Eq.\eqref{eq:problem3.2.4} and Eq.\eqref{eq:problem3.2.5} are given in following subsections. See Algorithm \ref{algorithm2: RTC-PFNC} for the pseudocode of RTC-PFNC.

\subsubsection{Update variable $\mathcal{L}$}
\label{RTC-PFNC: update L}
According to \textbf{Lemma.1} and Eq.\eqref{eq:problem3.2.3}, the variable $\mathcal{L}$ can be updated by
\begin{align}
    \mathcal{L} _{k}^{l+1} &=\underset{\mathbf{L}}{\mathrm{arg}\min}\Big\{  \alpha _kf\left( \mathbf{L}_{k\left( k \right)} \right) +\left< \mathbf{T}_{k\left( k \right)}^{l}, \mathbf{L}_{k\left( k \right)} \right>\nonumber\\
    &~~~+\frac{\rho _k}{2}\left\| \mathbf{L}_{k\left( k \right)}+\mathbf{E}_{k\left( k \right)}^{l}-\mathbf{M}_{\left( k \right)}^{l} \right\| _{F}^{2}\Big\}\nonumber\\
    &=\underset{\mathbf{L}}{\mathrm{arg}\min}\Big\{\frac{\alpha _k}{\rho _k}f\left( \mathbf{L}_{k\left( k \right)} \right) \nonumber\\
    &~~~+\frac{1}{2}\left\| \mathbf{L}_{k\left( k \right)}-\left( \mathbf{M}_{\left( k \right)}^{l}-\mathbf{E}_{k\left( k \right)}^{l}-\frac{1}{\rho _k}\mathbf{T}_{k\left( k \right)}^{l} \right) \right\| _{F}^{2} \Big\}\nonumber\\
    &=\mathrm{fold}_k\left( \mathcal{D} _{\omega ^l,\tau}\left( \mathbf{M}_{\left( k \right)}^{l}-\mathbf{E}_{k\left( k \right)}^{l}-\frac{1}{\rho _k}\mathbf{T}_{k\left( k \right)}^{l} \right) \right), 
    \label{eq:problem3.2.7}
\end{align}

\subsubsection{Update variable $\mathcal{M}$} Similar to Eq.\eqref{eq:problem3.1.12}, the closed-form solution of the  $\mathcal{M}$ subproblem in Eq.\eqref{eq:problem3.2.4} is obtained as
\begin{align}
    \mathcal{M} ^{l+1}
    &=\frac{1}{\sum\nolimits_{k=1}^3{\rho _k}}\sum\nolimits_{k=1}^3{\left( \rho _k\mathcal{L} _{k}^{l+1}+\rho _k\mathcal{E} _{k}^{l}+\mathcal{T} _{k}^{l} \right)},
    \label{eq:problem3.2.8}
\end{align}

\subsubsection{Update variable $\mathcal{E}$}
The closed-form solution of the  $\mathcal{E}$ subproblem in Eq.\eqref{eq:problem3.2.5} is given by
\begin{align}
    \mathcal{E} _{k}^{l+1} 
    &=\underset{\mathcal{E} _k}{\mathrm{arg}\min}\Big\{ \lambda _k\left\| \mathcal{E} _k \right\| _1\nonumber\\
    &~~~+\frac{\rho _k}{2}\left\| \mathcal{E} _k-\left( \mathcal{M} ^{l+1}-\mathcal{L} _{k}^{l+1}-\frac{1}{\rho _k}\mathcal{T} _k \right) \right\| _{\mathrm{F}}^{2} \Big\} ~~~~\nonumber\\
    &=sgn\left( \mathcal{H} \right) \circ \max \left\{ \left| \mathcal{H} \right|-\frac{\lambda _k}{\rho _k},\,\,0 \right\}, ~~~~
    \label{eq:problem3.2.9}
\end{align}
where $\mathcal{H} =\mathcal{M} ^{l+1}-\mathcal{L} _{k}^{l+1}-\frac{1}{\rho _k}\mathcal{T} _k$,  $\circ$ indicates the pointwise product, and the $sgn\left( \cdot \right) $ denotes the signum function, i.e.,

\begin{align}
sgn\left( x \right) =\begin{cases}
	1  ~~  &\mathrm{if} ~ x>0\\
	0  ~~ &\mathrm{if} ~ x=0\\
	-1 ~~ &\mathrm{if} ~ x<0.\\
\end{cases}~~~~~~~~~~~~~~~~~~~~
\label{eq:problem3.2.10}
\end{align}

\begin{algorithm}
    \caption{Numerical solution of \eqref{eq:problem3.2.2} via ADMM}
    \KwIn{The Observed Data Tensor $\mathcal{Y}$}
    \KwOut{The Recovered Low-rank Tensor $\mathcal{L}$ and Anomaly Tensor $\mathcal{E}$}
    \textbf{Initialization:} $\mathcal{L} ^0=\mathcal{M}^0 =\mathcal{Y}, ~\mathcal{L}_k ^0=\mathcal{E}_k ^0=\mathcal{T}_k ^0=0, ~\alpha _1=\alpha _2=\alpha _3=\frac{1}{3}, ~\lambda_1 = \lambda_2 = \lambda_3, ~\rho_k=\rho =\rho^0, ~\varepsilon = 1\mathrm{e}-6$ ;
    \LinesNumbered
    
    \SetAlgoVlined
    \While{not converged}{
    \For{k=1:3}{
        Update $\mathcal{L} _{k}^{l+1}$ via Eq.\eqref{eq:problem3.2.7} \;
        }
    Update $\mathcal{L} ^{l+1}$ via Eq.\eqref{eq:problem3.1.11}\;
    Update $\mathcal{M} _{k}^{l+1}$ via Eq.\eqref{eq:problem3.2.8}\;
    \For{k=1:3}{
        Update $\mathcal{E} _{k}^{l+1}$ via Eq. \eqref{eq:problem3.2.9}\;
        }  
    Update $\tilde{\mathcal{T}}^{l+1}$ via Eq.\eqref{eq:problem3.2.6}\;
    
    $l=l+1$
    }
    \label{algorithm2: RTC-PFNC}
\end{algorithm}

\section{Experiment}
\label{sec:Experiment}

\subsection {Traffic Data Sets}

We used the following four spatiotemporal traffic data sets for our experiment.
\begin{itemize}
\item(\textbf{P}): PeMS freeway traffic volume data set. This data set contains traffic volume
collected from 228 loop detectors with a 5-minute resolution (i.e., 288 time intervals per day)  over the weekdays of May and June, 2012 in District 7 of California by Caltrans Performance Measurement System (PeMS). The tensor size is 228 × 288 × 44.
\item(\textbf{S}): Seattle freeway traffic speed data set. This data set contains freeway traffic speed from 323 loop detectors with a 5-minute resolution (i.e., 288 time intervals per day) over the first four weeks of January, 2015 in Seattle, USA. 
The tensor size is 323 × 288 × 28. 
\item(\textbf{G}): Guangzhou urban traffic speed data set. This data set contains traffic speed collected from 214 road segments over two months (from August 1 to September 30, 2016) with a 10-minute resolution (i.e., 144 time intervals per day) in Guangzhou, China. The tensor size is 214 × 144 × 61.
\item(\textbf{B}): Birmingham parking occupancy data set. This data set registers occupancy (i.e., number of parked vehicles) of 30 car parks in Birmingham City for every half an hour between 8:00 and 17:00 over more than two months (77 days from October 4, 2016 to December 19, 2016). The tensor size is 30 × 18 × 77.
\end{itemize}

\subsection {Missing Data Generation}
\label{Missing Data Generation}
To test the missing data recovery capability of the proposed method, we configured two data missing patterns: random missing (RM), and non-random missing (NM). 
RM and NM data were generated by referring to Chen et al \cite{chen2020nonconvex}. 
According to the mechanism of RM and NM data, we masked a certain amount of observations as missing values (e.g., 20$\%$, 40 $\%$, 60$\%$, 80$\%$), and the remaining partial observations as input data for learning a well-behaved model. The experimental results of missing data recovery with respect to the two missing patterns were presented and analyzed in Section \ref{Experimental Results on Missing Data}.
  
\subsection {Corrupted and Missing Data Generation}
\label{Corrupted Missing Data Generation}
Considering both the missing data recovery accuracy and robustness to outlier corruption, on the basis of two data missing patterns, we sampled sparse outliers and randomly added them to the partial observed entries. The corrupted and missing data entries $y$ of observed tensor $\mathcal{Y}$ were generated as follows
 \begin{align}
    y=\,\,\begin{cases}
        \left[ y_i+\epsilon \right] _+\,\,&		y_i\in \Omega _c\\
        y_i\,\,&		y_i\in \Omega _o\,\\
        0&		y_i\in \Omega ^{\bot},\\
    \end{cases}
 \label{eq:corruption}
 \end{align}
where $\epsilon $ indicates the sparse outlier randomly sampled from a uniform distribution $\mathrm{U}\left( -s, s \right)$, $s$ represent the maximum magnitude of the added outliers, $\left[ \cdot \right] _+$ is an operator to ensure that the data entries are positive, $\Omega _n$ denotes the sparse outlier position set randomly selected from observation set $\Omega$ with a fraction of $\gamma $, $\Omega _c\cup \Omega _o=\Omega$. The experimental results of corrupted and missing data recovery were displayed in Section \ref{Experimental Results on Corrupted and Missing Data}.

\subsection {Evaluation Metrics}
\label{Evaluation Metrics}
To assess the data recovery performance, we used the actual values (ground truth) of these entries to compute the metrics MAPE and RMSE
 \begin{align}
    &\mathrm{MAPE}=\frac{1}{n}\sum\nolimits_{i=1}^n{\left| \frac{y_i-\hat{y}_i}{y_i} \right|\times 100},\nonumber\\
    &\mathrm{RMSE}=\sqrt{\frac{1}{n}\sum\nolimits_{i=1}^n{\begin{array}{c}
        \left( y_i-\hat{y}_i \right) ^2\\
    \end{array}}},
 \label{eq:MAPE & RMSE}
 \end{align}
where $y_i$ and $\hat{y}_i$ represent the actual value and recovered value of entry $i$ in the missing position, respectively.

\subsection {Baseline Models}\label{Baseline Models}
For comparison, we choose four baseline models from the following categories: Matrix Factorization, 
Tensor Factorization, Rank Minimization (RM) with convex relaxation, and RM with non-convex relaxation, respectively:
\begin{itemize}
    \item Bayesian Temporal Matrix Factorization (BTMF,\cite{chen2021bayesian}). This is a fully Bayesian matrix factorization model which integrates the vector autoregressive (VAR) model into the latent temporal factors. 
    \item Bayesian Gaussian CP decomposition (BGCP,\cite{chen2019bayesian}). This is a fully Bayesian tensor factorization model which uses Markov chain Monte Carlo to learn the latent factor matrices (i.e., low-rank structure).
    \item High-accuracy Low-Rank Tensor Completion (HaLRTC,\cite{liu2012tensor}). This is a LRTC model which uses nuclear norm minimization to find an accurate estimation of unobserved/missing entries in tensor data.
    \item Low-Rank Tensor Completion with Truncation Nuclear Norm minimization (LRTC-TNN,\cite{chen2020nonconvex}). This is a low-rank completion model in which nonconvex truncated nuclear norm minimization can help maintain the most important low-rank patterns. 
\end{itemize}

\subsection {Parameter Setting}
\label{Parameter Setting}
For TC-PFNC, there is no additional parameter except the convergence criterion, which is fixed as $abs\left( o_{l+1}-o_l \right) /o_l<1e-6$, where $o_{l+1}$ denotes the objective value at $\left( l+1 \right)$th iteration. For RTC-PFNC, only the parameter $\lambda$ is needed to be tuned. In Section \ref{The robustness to parameter selection}, we will show the robustness of RTC-PFNC to the selection of parameter $\lambda$. The convergence criterion of RTC-PFNC is also fixed as $abs\left( o_{l+1}-o_l \right) /o_l<1e-6$. The penalty parameter $\rho$ corresponds to the learning rate of the ADMM algorithm. The smaller the value of $\rho$ is, the faster the learning process of ADMM will be. In this study, we set $\rho$ value constant instead of a decay form to avoid introducing more hyperparameters. In all experiments, we select the parameters achieving the best performance for alternative models.

\section{Results}
\label{sec:Results}

\begin{table*}
    \centering
    \caption{Performance comparison (in MAPE/RMSE) among TC-PFNC and baseline models for RM and NM data recovery with varying missing rates.} \label{tab:performance comparison of missing data}
    \begin{tabular}{c|c|c|c|c|c||c} 
    \hline
    \hline
    \multirow{2}{*}{Data} & \multirow{2}{*}{Missing} & \multicolumn{2}{c|}{\begin{tabular}[c]{@{}c@{}}Matrix/Tensor Factorization \\Based Baseline\end{tabular}} & \multicolumn{2}{c||}{\begin{tabular}[c]{@{}c@{}}Rank Minimization \\Based Baseline\end{tabular}} & Proposed               \\ 
    \cline{3-7}
                          &                          & BTMF\cite{chen2021bayesian}                 & BGCP\cite{chen2019bayesian}                                                                              & HaLRTC\cite{liu2012tensor}       & LRTC-TNN\cite{chen2020nonconvex}                                                                        & TC-PFNC              \\ 
    \hline
    \multirow{8}{*}{P}    & 20\% RM                  & 6.96/4.73            & 9.34/6.27                                                                         & 4.20/2.81    & 3.19/2.26                                                                       & \textbf{2.33/1.73}     \\
                          & 40\% RM                  & 7.26/4.91            & 9.36/6.27                                                                         & 5.40/3.57    & 3.83/2.71                                                                       & \textbf{2.70/1.99}     \\
                          & 60\% RM                  & 7.66/5.16            & 9.43/6.30                                                                         & 7.01/4.53    & 4.77/3.37                                                                       & \textbf{3.35/2.46}     \\
                          & 80\% RM                  & 8.58/5.69            & 9.39/6.29                                                                         & 9.51/5.89    & 6.41/4.45                                                                       & \textbf{5.40/3.78}     \\
                          & 20\% NM                  & 9.10/6.14            & 9.32/6.36                                                                         & 8.54/5.56    & 6.93/4.96                                                                       & \textbf{6.79/4.84}     \\
                          & 40\% NM                  & 9.51/6.30            & 9.88/6.54                                                                         & 9.92/6.16    & 7.83/5.44                                                                       & \textbf{7.74/5.32}     \\
                          & 60\% NM                  & 9.97/6.51            & 10.21/6.69                                                                        & 11.54/6.89   & 8.77/5.97                                                                       & \textbf{8.68/5.82}     \\
                          & 80\% NM                  & 11.97/\textbf{7.92}           & 14.43/12.67                                                                       & 15.20/9.48   & 10.86/8.35                                                                      & \textbf{10.49}/7.96   \\ 
    \hline
    \multirow{8}{*}{S}    & 20\% RM                  & 5.92/3.71            & 7.46/4.50                                                                         & 5.93/3.47    & 4.65/3.06                                                                       & \textbf{4.47/2.96}     \\
                          & 40\% RM                  & 6.18/3.79            & 7.54/4.54                                                                         & 6.76/3.83    & 5.12/3.30                                                                       & \textbf{4.85/3.15}     \\
                          & 60\% RM                  & 6.38/3.95            & 7.60/4.56                                                                         & 7.90/4.34    & 5.80/3.66                                                                       & \textbf{5.35/3.41}     \\
                          & 80\% RM                  & 7.33/4.41            & 7.82/4.68                                                                         & 10.26/5.31   & 7.71/4.64                                                                       & \textbf{6.58/4.05}     \\
                          & 20\% NM                  & 9.08/5.30            & 9.90/5.67                                                                         & 8.69/4.74    & 7.06/4.27                                                                       & \textbf{6.15/3.82}     \\
                          & 40\% NM                  & 9.30/5.37            & 10.11/5.73                                                                        & 10.27/5.32   & 7.75/4.60                                                                       & \textbf{6.80/4.15}     \\
                          & 60\% NM                  & 9.68/5.57            & 10.29/5.83                                                                        & 12.58/6.18   & 8.59/5.06                                                                       & \textbf{7.99/4.75}     \\
                          & 80\% NM                  & 11.58/\textbf{6.52}           & 13.63/8.99                                                                        & 19.57/10.91  & 11.51/8.51                                                                      & \textbf{11.32}/8.42    \\ 
    \hline
    \multirow{8}{*}{G}    & 20\% RM                  & 7.46/3.19            & 8.30/3.58                                                                         & 8.13/3.33    & 6.70/2.88                                                                       & \textbf{6.48/2.76}     \\
    & 40\% RM                  & 7.81/3.35            & 8.34/3.60                                                                         & 8.86/3.61    & 7.32/3.17                                                                       & \textbf{7.11/3.02}     \\
    & 60\% RM                  & 8.51/3.65            & 8.45/3.65                                                                         & 9.82/3.96    & 8.12/3.51                                                                       & \textbf{7.93/3.36}     \\
    & 80\% RM                  & 9.80/4.21            & \textbf{8.76}/3.80                                                                & 11.32/4.48   & 9.25/3.98                                                                       & 8.90\textbf{/3.76}     \\
    & 20\% NM                  & 10.31/4.27            & 10.40/4.31                                                                        & 10.57/4.24   & \textbf{9.52}/4.01                                                                       & 9.53/\textbf{3.99}     \\
    & 40\% NM                  & 10.26/4.31            & 10.37/4.34                                                                        & 10.94/4.39    & \textbf{9.61}/4.09                                                                       & 9.65\textbf{/4.08}     \\
    & 60\% NM                  & 10.55/4.45           & 10.72/4.83                                                                        & 11.82/4.69   & \textbf{9.87}/\textbf{4.22}                                                     & 9.98\textbf{/4.22}     \\
    & 80\% NM                  & 11.50/4.91           & 13.05/6.42                                                                        & 14.28/5.61   & \textbf{10.55}/4.48                                                             & 10.66\textbf{/4.45}    \\ 
    \hline
    \multirow{8}{*}{B}    & 20\% RM                  & \textbf{2.03/12.36}  & 6.31/19.73                                                                        & 5.70/21.57   & 4.75/15.45                                                                      & 4.21/13.06             \\
                          & 40\% RM                  & \textbf{3.12/15.29}  & 5.95/20.68                                                                        & 7.43/32.01   & 5.56/19.75                                                                      & 4.80/16.51             \\
                          & 60\% RM                  & 8.04/30.99           & 6.73/23.31                                                                        & 11.14/55.02  & 7.62/27.15                                                                      & \textbf{6.25/22.49}    \\
                          & 80\% RM                  & 16.41/84.04          & 9.93/38.83                                                                        & 18.58/112.36 & 10.58/40.23                                                                     & \textbf{9.30/36.64}    \\
                          & 20\% NM                  & 9.63/\textbf{37.29}  & 9.35/57.02                                                                        & 9.38/73.01   & 8.09/48.93                                                                      & \textbf{7.56}/47.90    \\
                          & 40\% NM                  & 10.09/\textbf{42.85} & 12.73/103.31                                                                      & 13.96/163.52 & 10.53/61.18                                                                     & \textbf{9.07}/51.21    \\
                          & 60\% NM                  & 17.66/\textbf{95.39} & 21.26/156.49                                                                      & 23.35/339.32 & 16.65/108.91                                                                    & \textbf{14.69}/105.12  \\
                          & 80\% NM                  & 33.62/189.56         & 28.21/168.77                                                                      & 40.39/597.97 & 35.84/591.10                                                                    & \textbf{24.76/151.27}  \\
    \hline
    \hline
    \end{tabular}
\end{table*}

\begin{figure*}[h]
    \centering
    \subfigure[PeMS Freeway Traffic Volume Data Set (P)]{\label{fig:TC-P-link86}\includegraphics[width=1\textwidth]{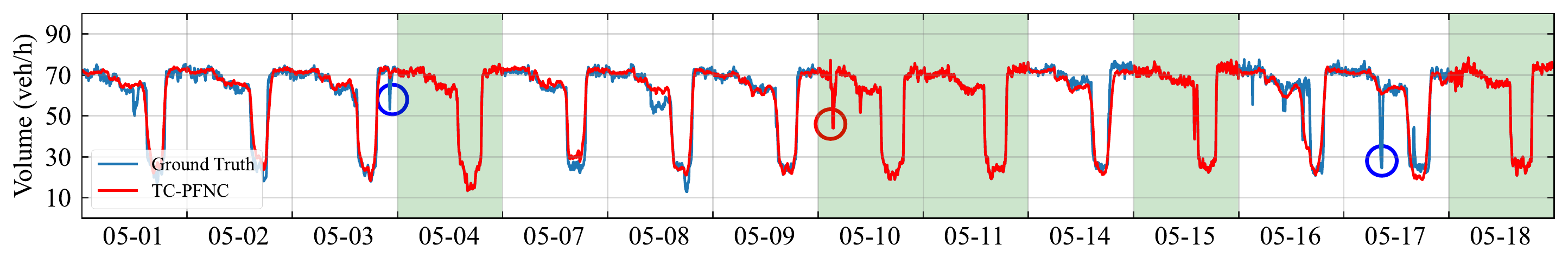}}
    \subfigure[Seattle Freeway Traffic Speed Data Set (S)]{\label{fig:TC-S-link9}\includegraphics[width=1\textwidth]{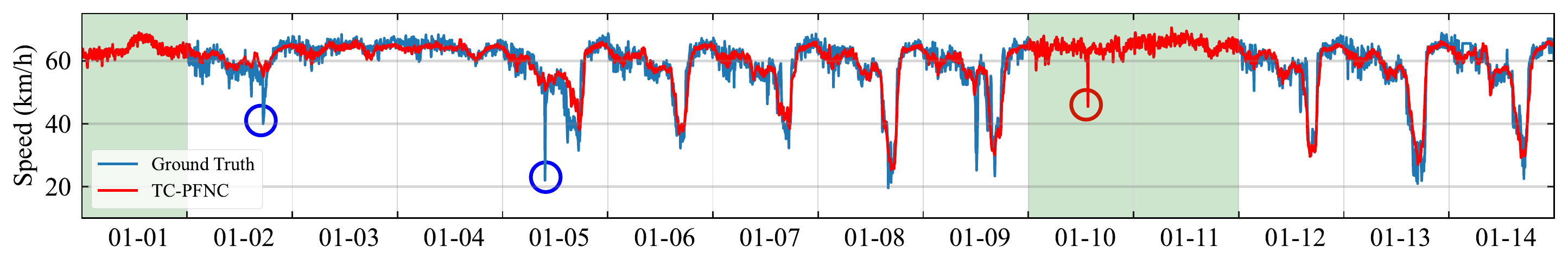}}
    \subfigure[Guangzhou Urban Traffic Speed Data Set (G)]{\label{fig:TC-G-link103}\includegraphics[width=1\textwidth]{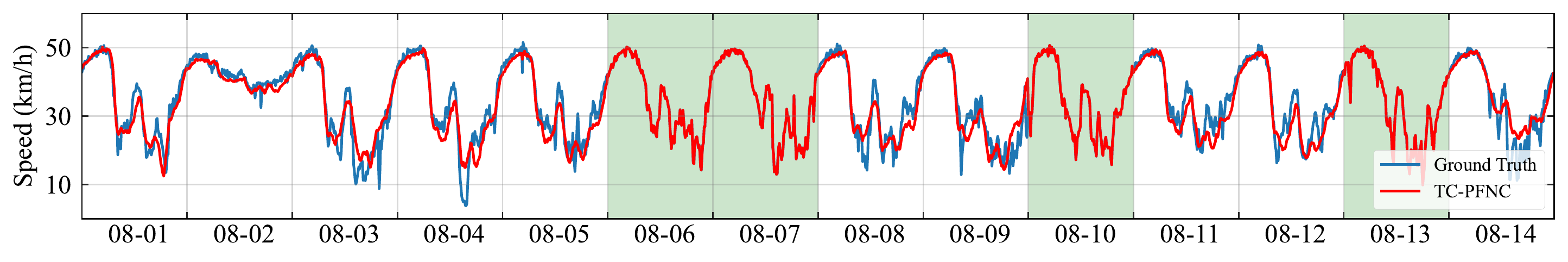}}\centerline{G}
    \subfigure[Birmingham Parking Occupancy Data Set (B)]{\label{fig:TC-B-link6}\includegraphics[width=1\textwidth]{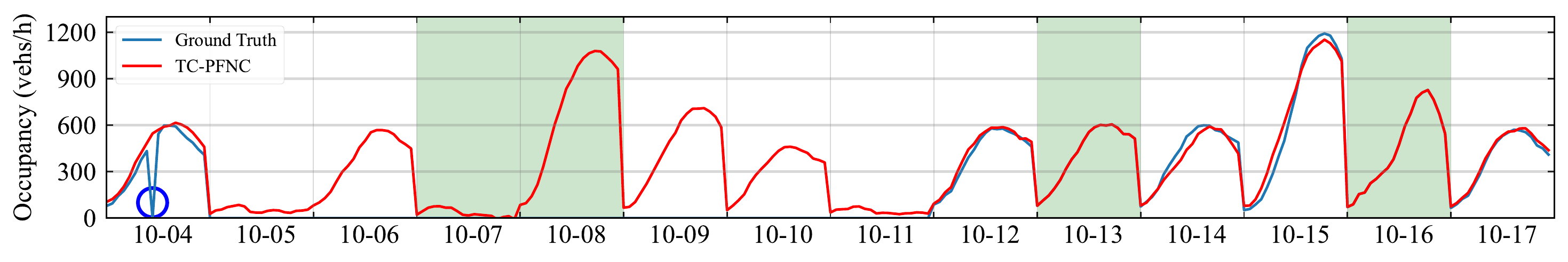}}
    \caption{Recovery example for four traffic data sets. In these panels, White rectangle represent fiber missing (i.e., observations are missing in a whole day), and green rectangles show the partially observed data.}
    \label{fig:TC-PFNC}
\end{figure*} 

\begin{figure*}[h]
    \centering

    \subfigure[Data Set (P)]{\label{fig:Corruption Performance-P} \includegraphics[width=0.3\textwidth]{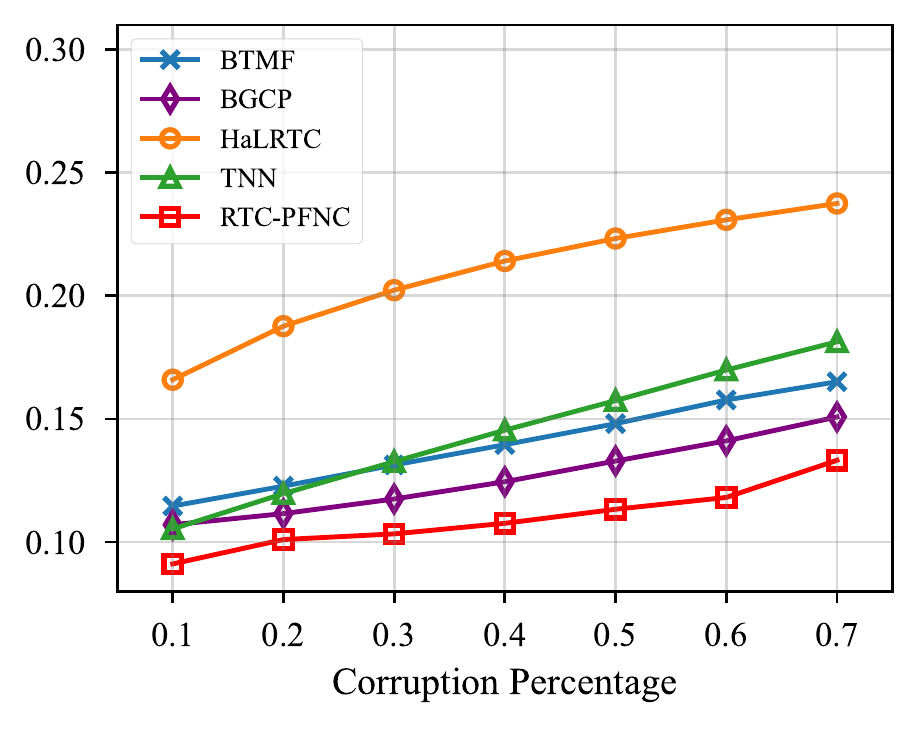}}
    \subfigure[Data Set (S)]{\label{fig:Corruption Performance-S} \includegraphics[width=0.3\textwidth]{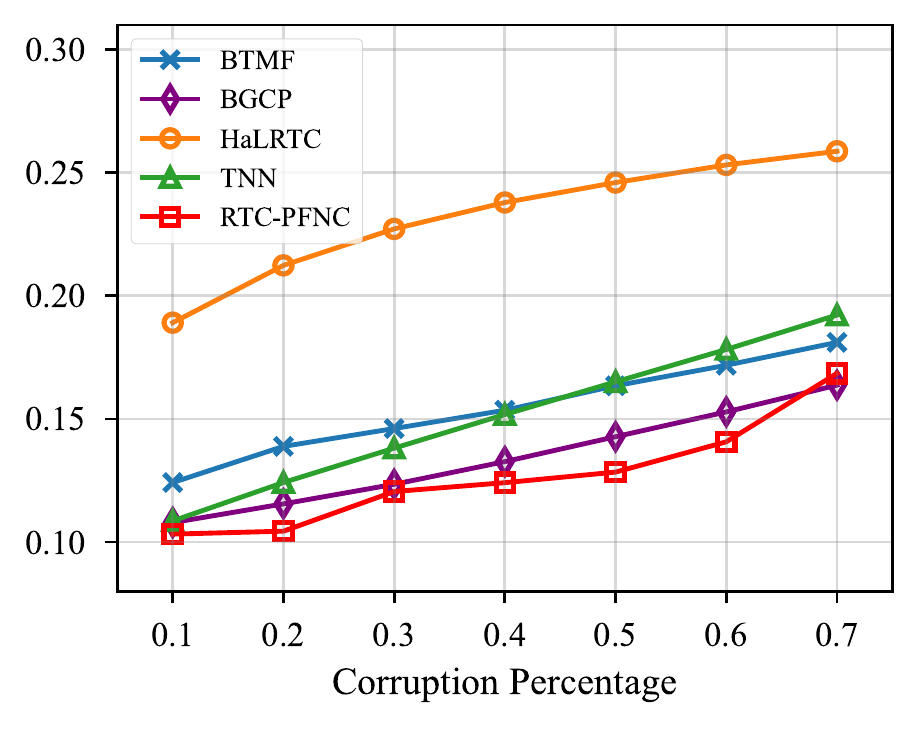}}
    \subfigure[Data Set (G)]{\label{fig:Corruption Performance-G} \includegraphics[width=0.3\textwidth]{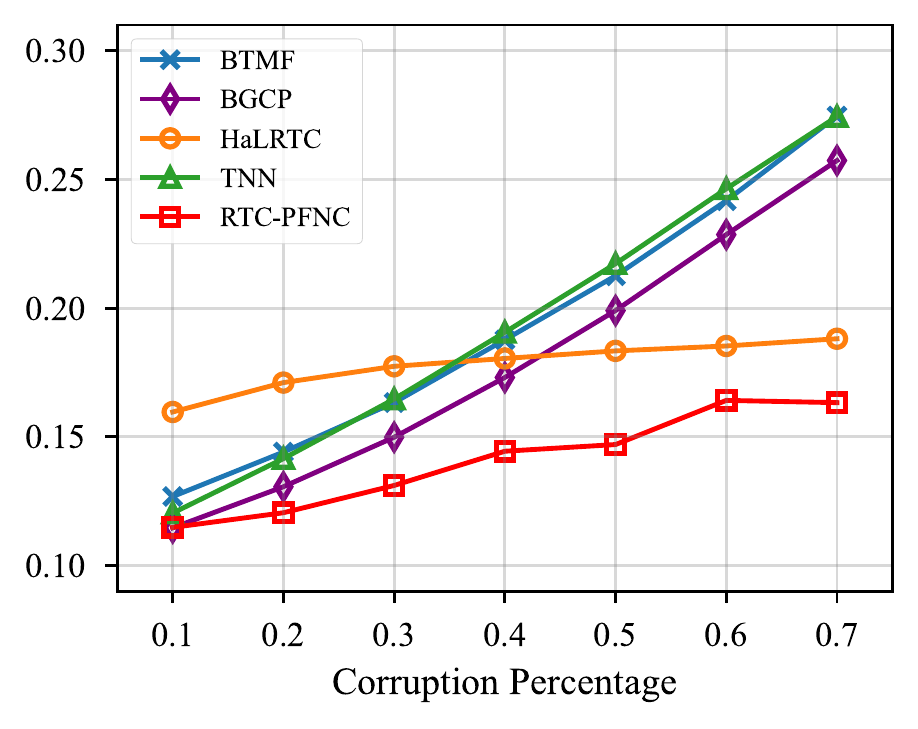}}
    \vspace{-0.25cm}
    \caption{The performance comparison (in MAPE) among RTC-PFNC and baseline models with respect to corruption percentage $\gamma$. 
    }
    \label{fig:Performance evolution}
\end{figure*} 
\begin{figure*}[h]
    \centering


    \subfigure[HaLRTC on (P)]{\label{fig:Performance Comparison-P-HaLRTC} \includegraphics[width=0.32\textwidth]{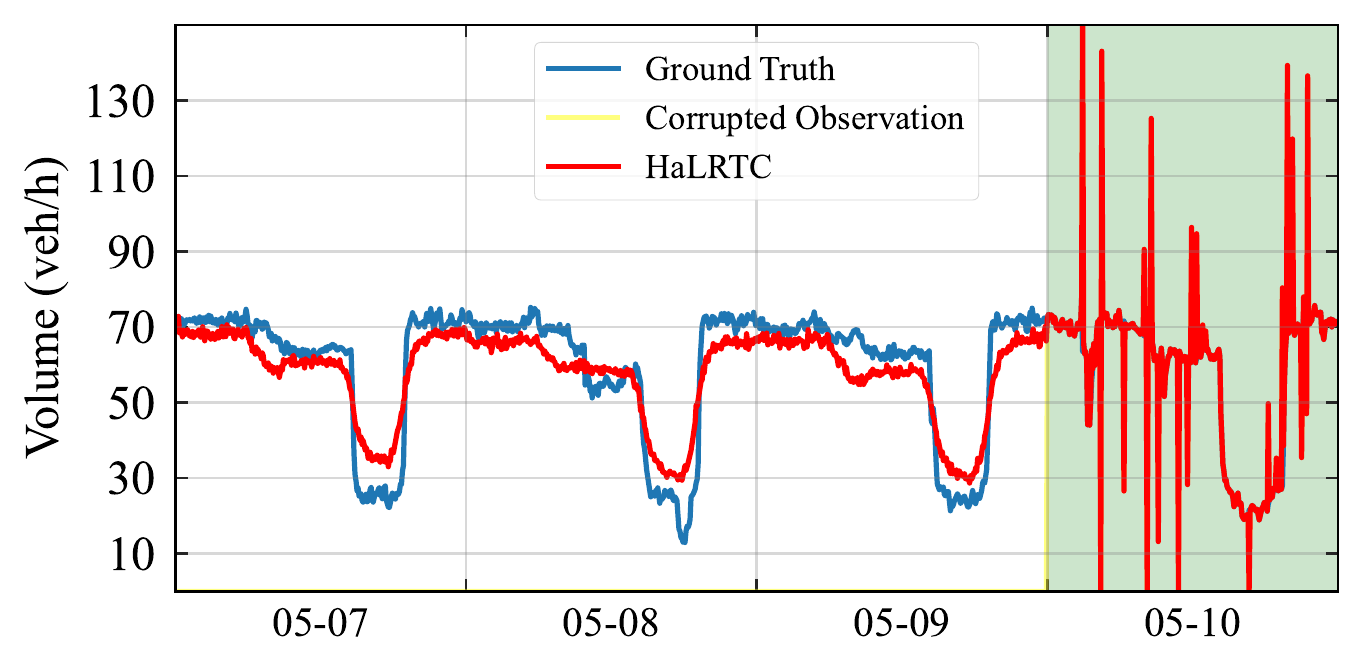}}
    \subfigure[HaLRTC on (S)]{\label{fig:Performance Comparison-S-HaLRTC} \includegraphics[width=0.32\textwidth]{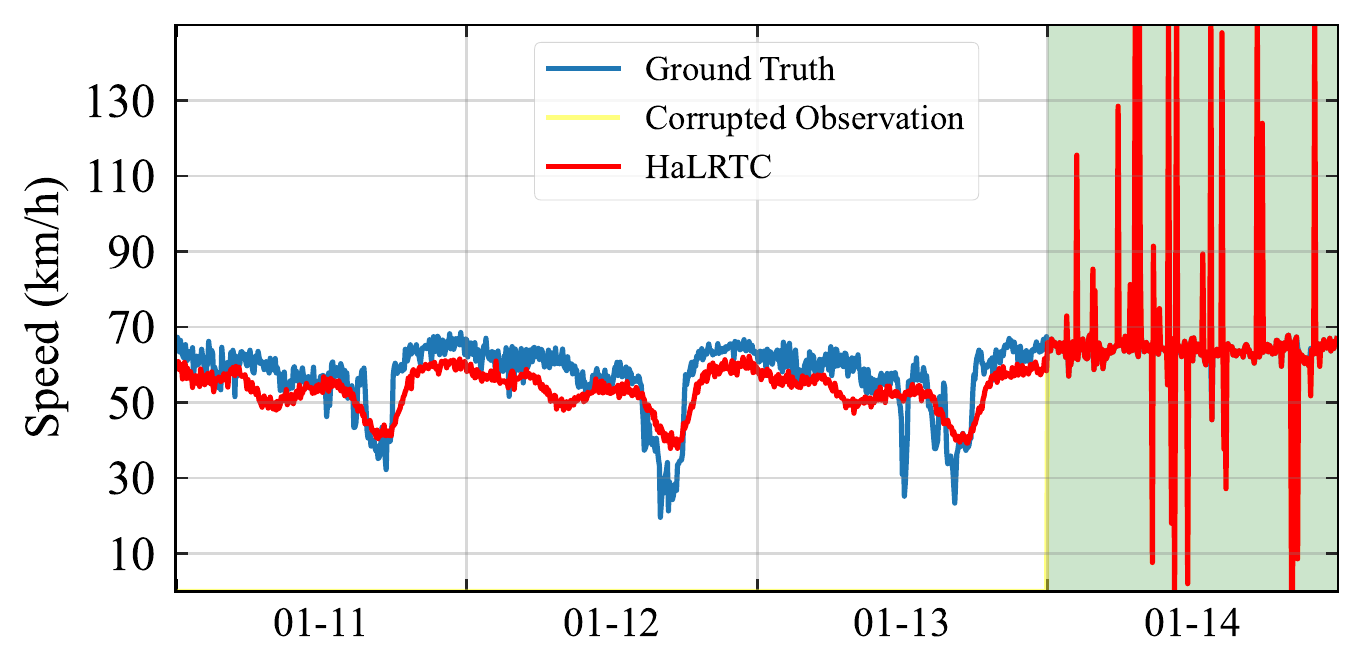}}
    \subfigure[HaLRTC on (G)]{\label{fig:Performance Comparison-G-HaLRTC} \includegraphics[width=0.32\textwidth]{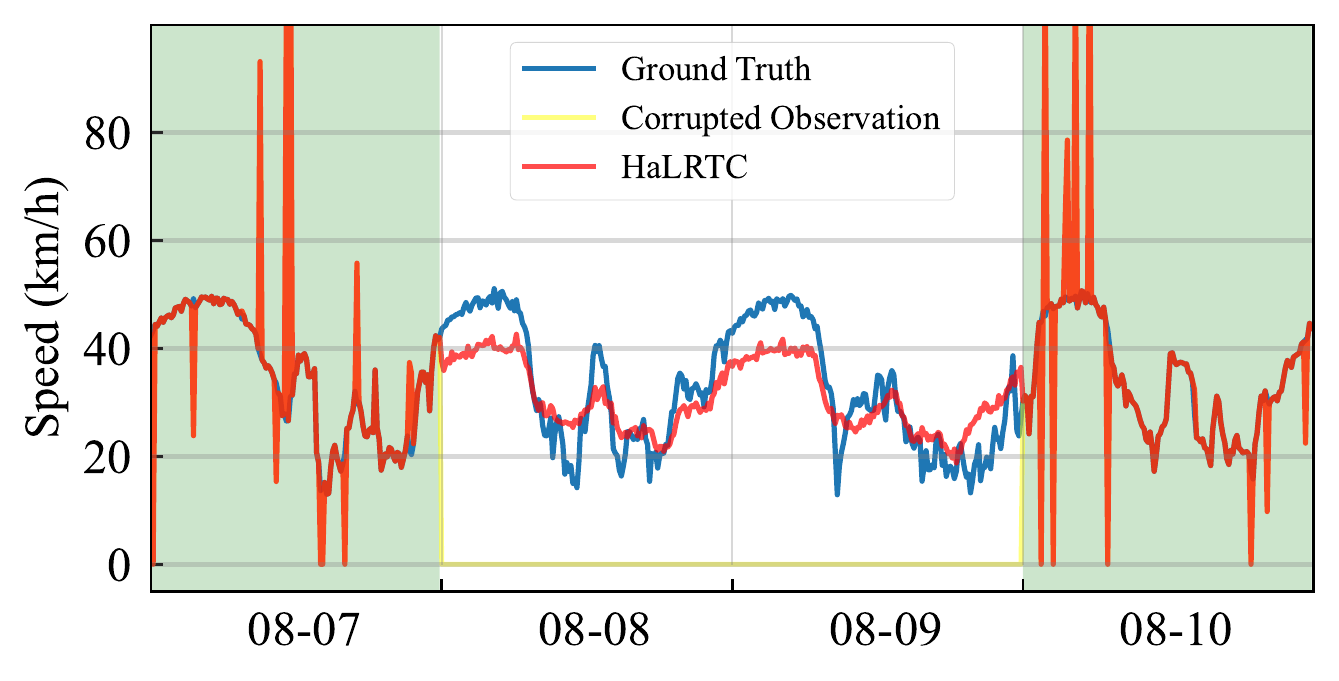}}

    \subfigure[LRTC-TNN on (P)]{\label{fig:Performance Comparison-P-LRTC-TNN} \includegraphics[width=0.32\textwidth]{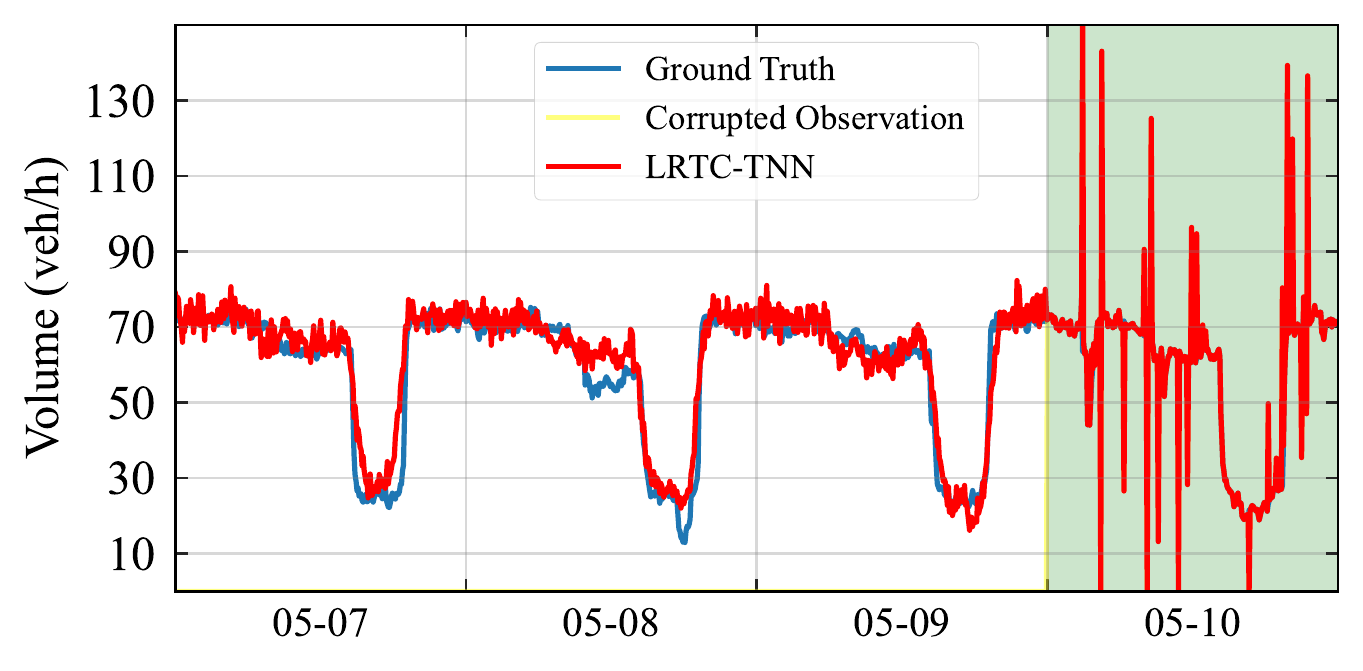}}
    \subfigure[LRTC-TNN on (S)]{\label{fig:Performance Comparison-S-LRTC-TNN} \includegraphics[width=0.32\textwidth]{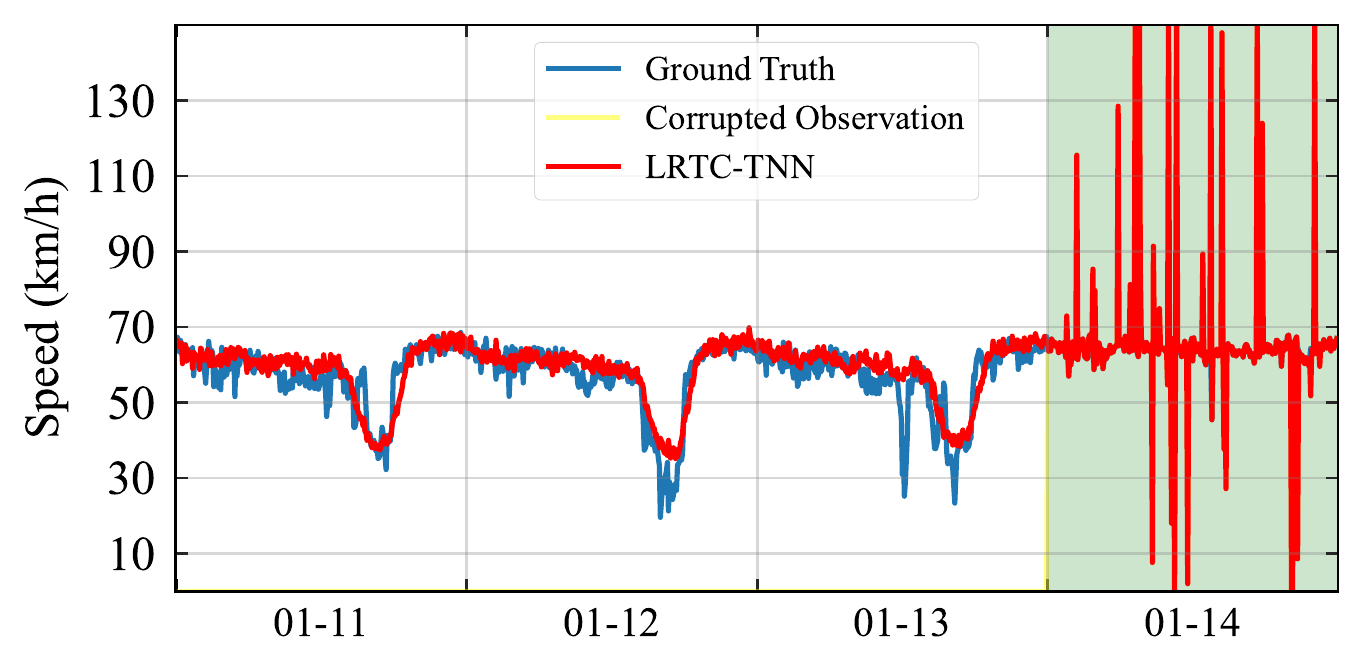}}
    \subfigure[LRTC-TNN on (G)]{\label{fig:Performance Comparison-G-LRTC-TNN} \includegraphics[width=0.32\textwidth]{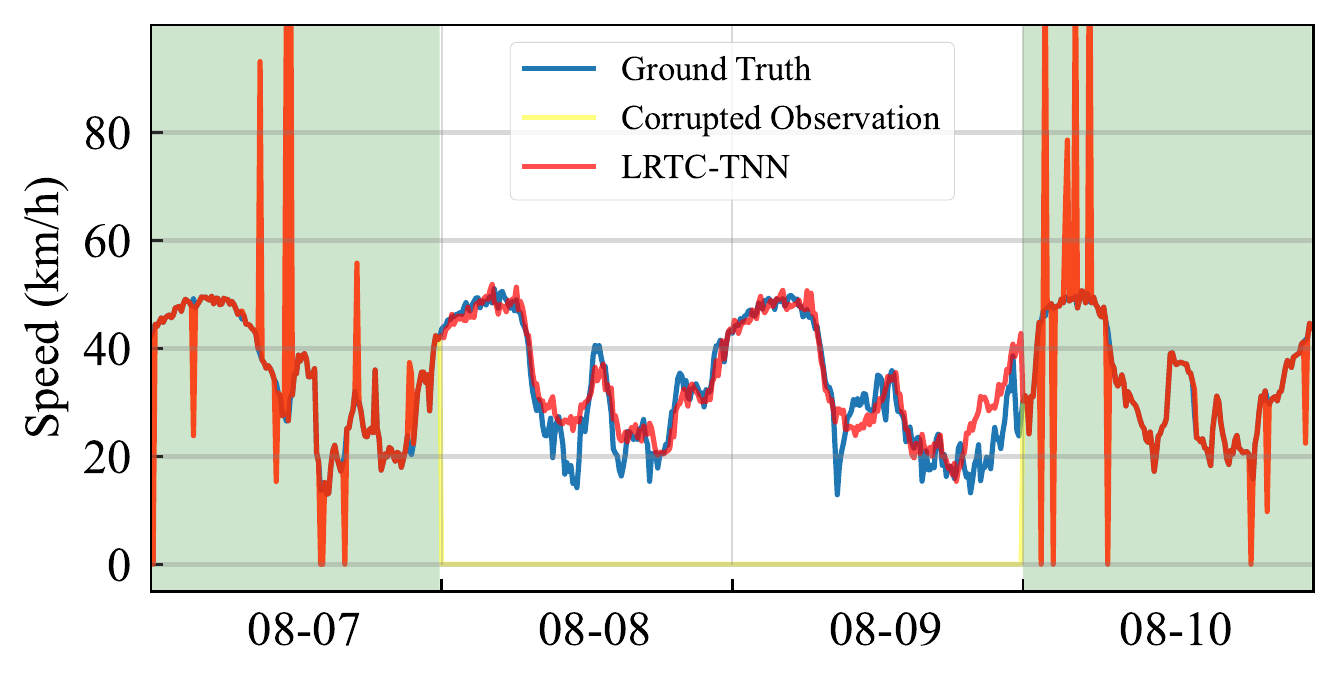}}

    \subfigure[Proposed RTC-PFNC on (P)]{\label{fig:Performance Comparison-P-PFNC} \includegraphics[width=0.32\textwidth]{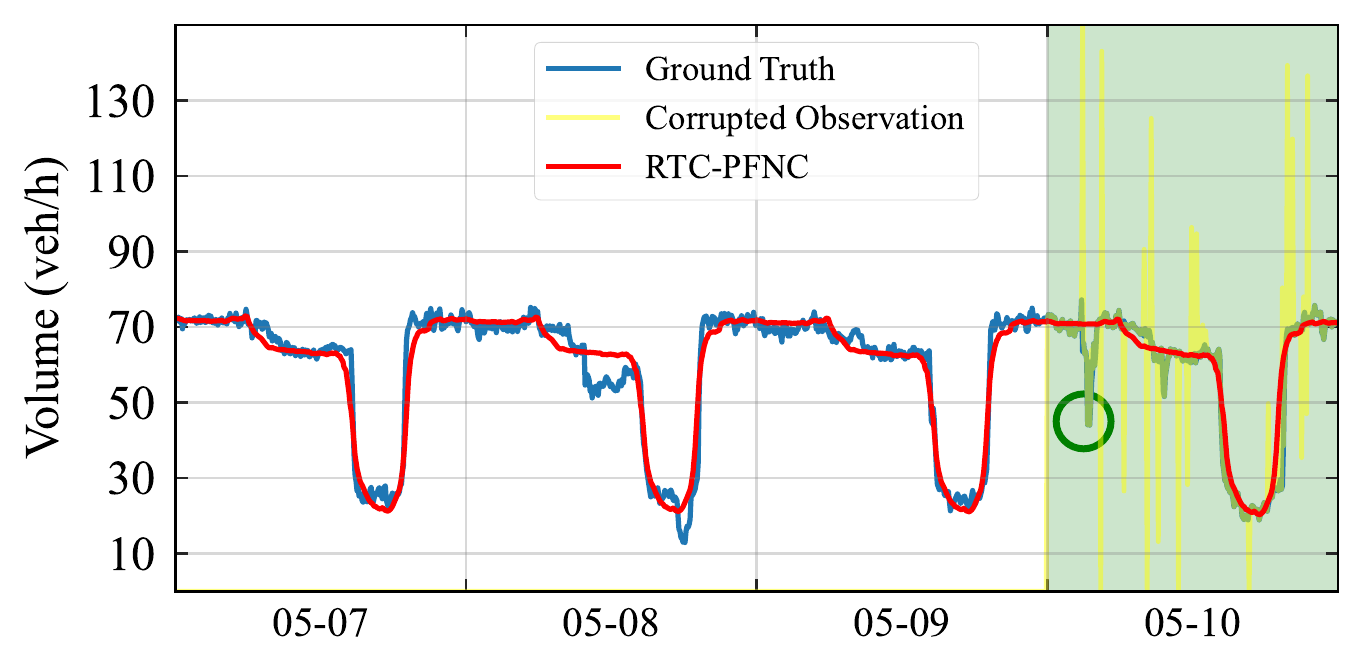}}
    \subfigure[Proposed RTC-PFNC on (S)]{\label{fig:Performance Comparison-S-PFNC} \includegraphics[width=0.32\textwidth]{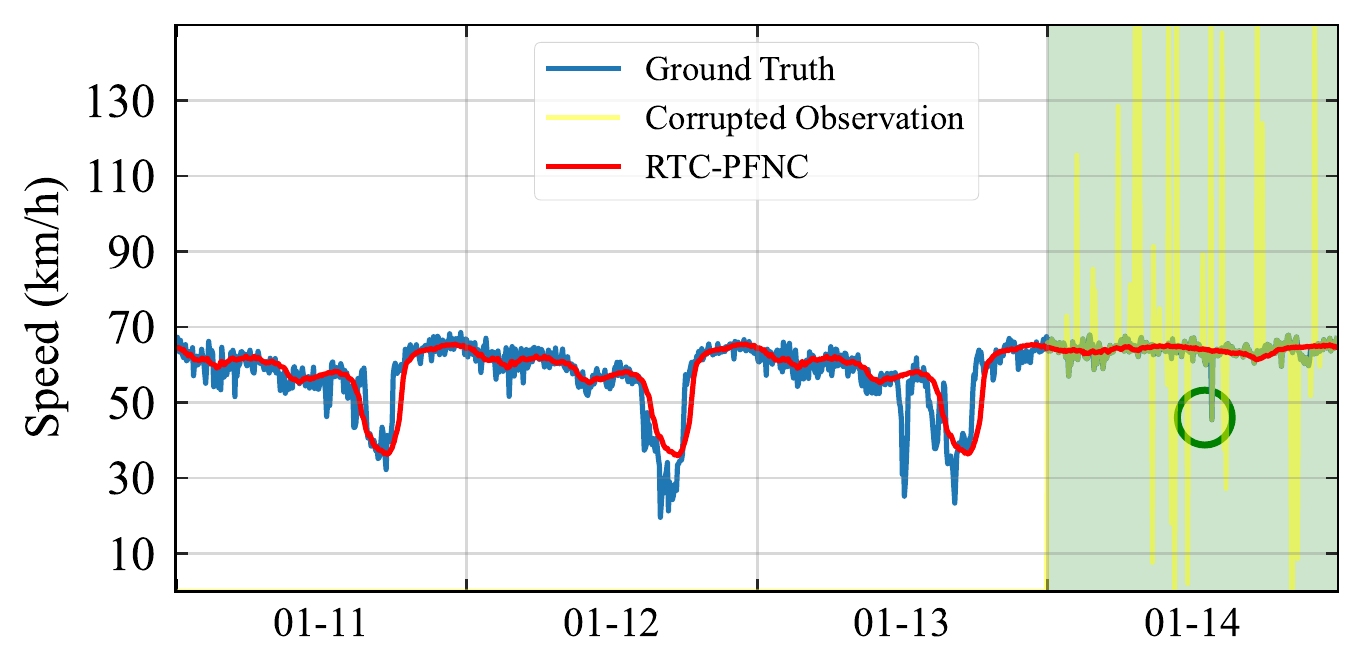}}
    \subfigure[Proposed RTC-PFNC on (G)]{\label{fig:Performance Comparison-G-PFNC} \includegraphics[width=0.32\textwidth]{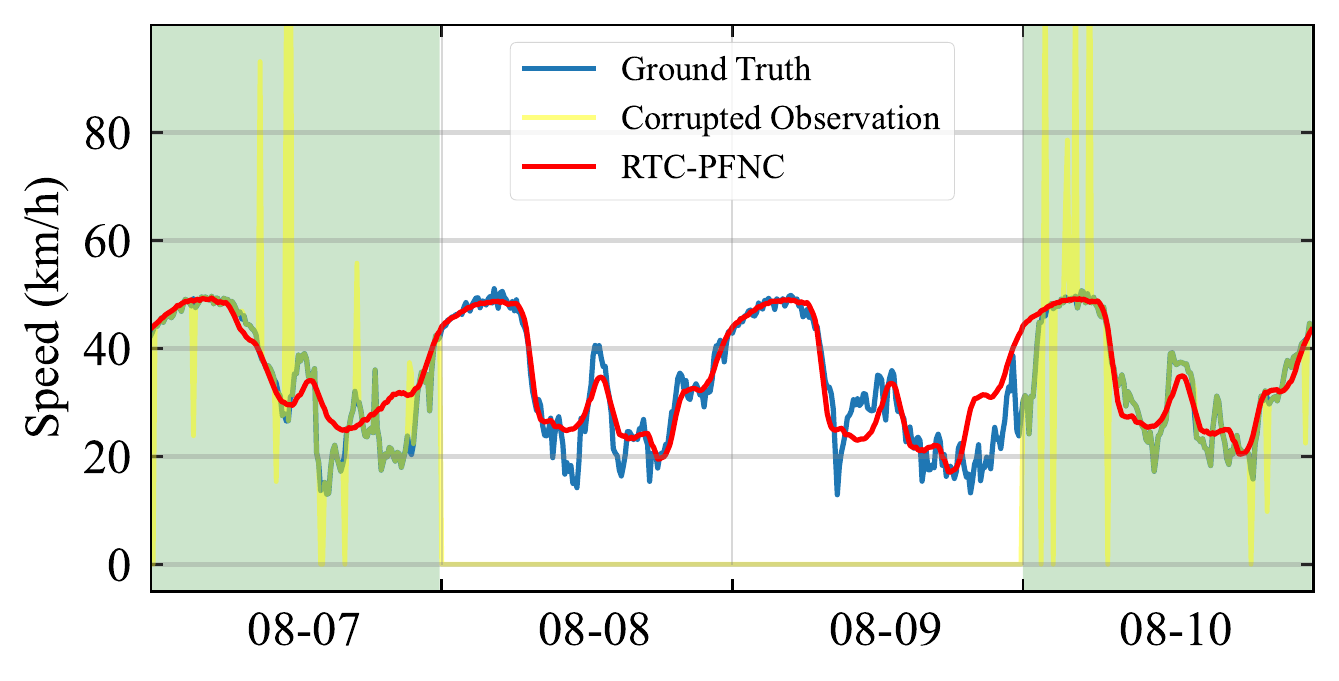}}
    \caption{The corrupted and non-random missing data recovery examples of proposed RTC-PFNC and baseline low-rank tensor completion (LRTC) models on data set (G), (P), and (S) (corruption percentage $\gamma=0.1$) . 
    }
    \label{fig:Corruption Performance Comparison}
\end{figure*} 
\subsection {Experimental Results on Missing Data}
\label{Experimental Results on Missing Data}

Table \ref{tab:performance comparison of missing data} summarizes the recovery performance of TC-PFNC and four representative baselines selected from matrix factorization (BTMF\cite{chen2021bayesian}), tensor factorization (BGCP\cite{chen2019bayesian}), rank minimization with convex relaxation (HaLRTC\cite{liu2012tensor}), and rank minimization with nonconvex relaxation (LRTC-TNN\cite{chen2020nonconvex}) on the four selected traffic data sets with various missing scenarios. Of these results, the NM data seems to be more difficult to reconstruct with all these recovery models than the RM data.

Overall, both the LRTC-TNN and proposed TC-PFNC, the nonconvex relaxation-based LRTC models, achieved promising performance with lower MAPE/RMSE values than other baseline models.  Note that the performance of LRTC-TNN heavily depends on the additional parameter truncation rate $r$. A typical higher truncation rate $r$ (e.g., 0.3) was used for LRTC-TNN in RM scenarios, and a lower truncation rate $r$ (e.g., 0.05) was preferred in NM scenarios. As for TC-PFNC, there is no parameter needed to be tuned. Comparing TC-PFNC with LRTC-TNN, the result showed the advantages of the proposed parameter-free nonconvex relaxation, i.e., log-based relaxation provides higher traffic data recovery accuracy.

For data set (P), (S), and (G), the proposed TC-PFNC outperformed other baselines in diverse missing scenarios (RM and NM scenarios with varying missing rates). For the Birmingham(B) parking data set, TC-PFNC achieved promising recovery performance in the NM scenarios and RM scenarios with high missing rate(e.g., 60$\%$, 80$\%$). The Bayesian temporal matrix decomposition model BTMF performed best for the RM with low missing rate scenarios (i.e. 20$\%$, 40$\%$). This is due to the strong temporal patterns and consistency in data set (B), thus the temporal smoothness and dynamics play a more important role in recovering the actual value.

In Fig. \ref{fig:TC-PFNC}, we chose some examples from four experiment data sets and visualized the NM time series with an extreme missing rate of 60$\%$ and corresponding recovered time series by employing TC-PFNC. We found out that TC-PFNC can achieve very high accuracy for all four data sets with only 40$\%$ input. 
The recovered time series provided by TC-PFNC generally recovered the structure of missing traffic data and would not over-fit the local value (the \textcolor{blue}{blue} markup). As mentioned in Section \ref{Evaluation Metrics}, the ground truth is the raw actual values in each data set. It may contain a little fraction of abnormal values. Take the day 05-17 in Fig. \ref{fig:TC-P-link86} as an example, according to the traffic flow continuity, there is less likely that the traffic speed on freeway sharply drops from 60km/h to 30km/h and back to 60km/h within 5min, thus we considered it as an abnormal outlier. 

In the meantime, under the noiseless assumption, the recovered values in the observed set $\Omega$ were consistent with observed entries, even for the abnormal one (the \textcolor{red}{red} markup). Then the paper show the robust recovery performance in abnormal values detection and removal achieved by the proposed RTC-PFNC in Section \ref{Experimental Results on Corrupted and Missing Data}.

\subsection {Experimental Results on Corrupted and Missing Data}
\label{Experimental Results on Corrupted and Missing Data}
In this section, we chose the non-random missing scenarios of data set (P), (S), and (G) to test the performance of proposed RTC-PFNC and other low-rank tensor completion (LRTC) baseline models on data with only 40$\%$ entries and random corruption. We fixed the maximum corruption magnitude as 100, which is around the maximum value of three traffic data sets. Fig. \ref{fig:Performance evolution} summarizes the recovery performance evolution (MAPE) of RTC-PFNC and baseline models with respect to corruption percentage. 

For the PeMS freeway volume data set(P), RTC-PFNC outperforms all alternative methods under various corruption percentages. For the Seattle freeway volume data set(S), RTC-PFNC performed best in slight and medium corruption scenarios, the common conditions in real world. The MAPE of RTC-PFNC is still less than that of the nonconvex model LRTC-TNN at the extreme condition ($\gamma=0.7$). For the Guangzhou Urban traffic speed data set (G), the proposed RTC-PFNC presented the best performance in various corruption percentages among all alternatives. Additionally, RTC-PFNC performed stably even in the scenarios with high corruption percentages, while other baseline models degraded sharply.

Fig. \ref{fig:Corruption Performance Comparison} presents the recovery examples of HaLRTC, LRTC-TNN and the proposed RTC-PFNC. The results demonstrated RTC-PFNC's missing values recovery accuracy and robustness to corrupted observations. Compared with the baseline models, RTC-PFNC provided a smoother recovered traffic data time series, which corresponds to a lower rank recovered tensor. Compared with TC-PFNC in Fig.\ref{fig:TC-PFNC}, RTC-PFNC was also robust to the outlier in observations (the \textcolor{green}{green} markup).

\subsection {The Robustness to Parameter Selection}
\label{The robustness to parameter selection}
\begin{figure}[h]
    \begin{center}
    \includegraphics[width=2.5in]{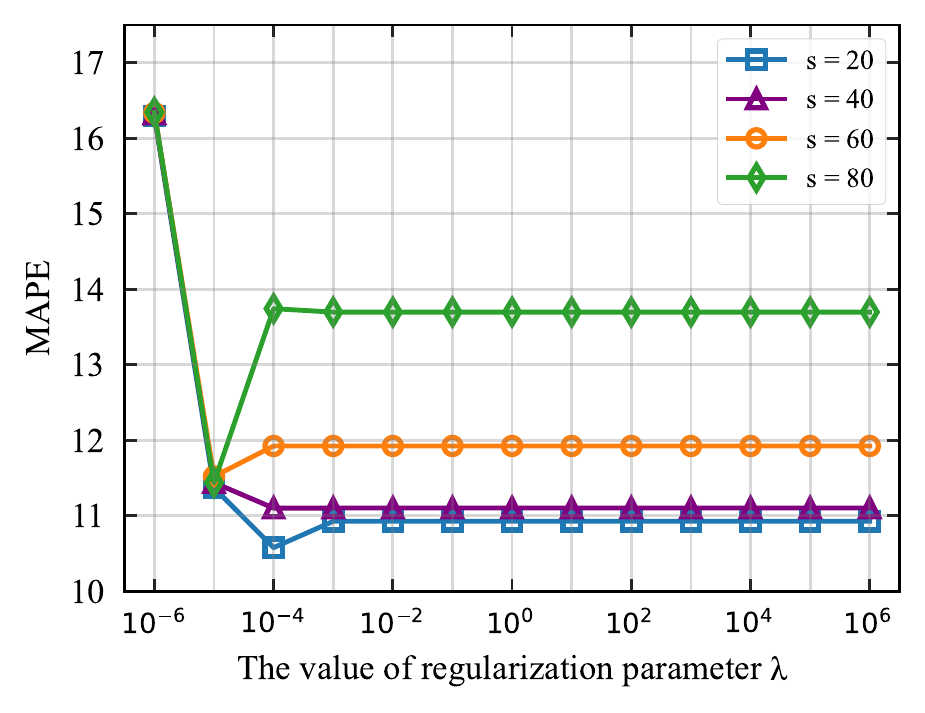}\\
    \caption{The experimental results of RTC-PFNC on Guangzhou urban traffic speed data set (G). $s$ denotes the maximum corruption magnitude defined in Section \ref{Missing Data Generation}. }\label{fig:Sensitivity analysis}
    \end{center}
\end{figure}

As mentioned in Section \ref{Parameter Setting}, only the parameter $\lambda$ was required to be determined in RTC-PFNC. To analyze the sensitivity of $\lambda$, we configured the value of $\lambda$ in the set $\{1e-6, 1e-5,..., 1e+6\}$. We selected only Guangzhou urban traffic speed data set (G) to serve as the test data set, and fixed the missing rate $ms =60\%$, corruption percentage $\gamma=0.1$. The experimental results of RTC-PFNC with various $\lambda$ were displayed in Fig. \ref{fig:Sensitivity analysis}, demonstrating the robustness of RTC-PFNC to the selection of parameter $\lambda$.

\section{Conclusion}
\label{sec:Conclusion}
In this paper, we first proposed a parameter-free nonconvex relaxation-based low-rank tensor completion model (TC-PFNC) to recover traffic data from partial observations. Then, considering the potential outliers in traffic data, we extended it to a robust version (RTC-PFNC) to recover the data from partial and corrupted observations and remove the anomalies in observations. Unlike the existing nonconvex models, both TC-PFNC and RTC-PFNC can simultaneously increase penalty to noise and decrease penalty to structural information without any parameter to calibrate, which enhances its applicability in real practice. We performed numerical experiments on real-world traffic data sets and the result demonstrated the significant superiority of PFNC.

There are several directions to advance this research. First, the spatial and temporal prior information (e.g. road segment similarity and temporal traffic flow consistency) is meaningful in real-world practice. A potential direction is to introduce spatial or temporal constraints to further improve the recovery accuracy. Second, only common missing patterns (i.e. RM and NM), were considered in the study. The proposed PFNC model can be modified to tackle more real-world missing scenarios, such as block-missing\cite{chen2021low} and all-missing (referred as to the Kriging problem in \cite{lei2022bayesian}).

\bibliographystyle{IEEEtran}
\bibliography{TITS}

\end{document}